\documentclass{article}

\usepackage{microtype}
\usepackage{graphicx}
\usepackage{subcaption}
\usepackage{booktabs} 
\usepackage{mathrsfs}
\usepackage{hyperref}

\usepackage[accepted]{icml2025}

\usepackage{amsmath}
\usepackage{amssymb}
\usepackage{mathtools}
\usepackage{amsthm}
\usepackage{multirow}
\definecolor{ForestGreen}{RGB}{34,139,34}
\usepackage[capitalize,noabbrev]{cleveref}
\usepackage[normalem]{ulem}
\usepackage[inline]{enumitem}
\usepackage[abs]{overpic}

\theoremstyle{plain}

\theoremstyle{definition}

\theoremstyle{remark}

\usepackage[textsize=tiny]{todonotes}

\icmltitlerunning{Upcycling Text-to-Image Diffusion Models for Multi-Task Capabilities}

\newcommand{\ie}{i.e.}

\newcommand{\expert}{E}

\newcommand{\Real}{\mathbb{R}}

\newcommand{\deW}{\Phi^{\tau}}
\newcommand{\convW}{\Psi_{C}}

\begin{document}

\twocolumn[
\icmltitle{Upcycling Text-to-Image Diffusion Models for Multi-Task Capabilities}



\icmlsetsymbol{equal}{*}

\begin{icmlauthorlist}
\icmlauthor{Ruchika Chavhan}{yyy}
\icmlauthor{Abhinav Mehrotra}{yyy}
\icmlauthor{Malcolm Chadwick}{equal,yyy}
\icmlauthor{Alberto Gil Ramos}{equal,yyy}
\icmlauthor{Luca Morreale}{equal,yyy}
\icmlauthor{Mehdi Noroozi}{equal,yyy}
\icmlauthor{Sourav Bhattacharya}{yyy}
\end{icmlauthorlist}

\icmlaffiliation{yyy}{Samsung AI Center, Cambridge, UK}
\icmlcorrespondingauthor{Ruchika Chavhan}{r2.chavan@samsung.com}

\icmlkeywords{Machine Learning, ICML}

\vskip 0.3in
]



\printAffiliationsAndNotice{\icmlEqualContribution} 

\begin{figure*}[t!]
\captionsetup[subfigure]{labelformat=empty}
    \centering
\begin{subcaptionbox}{}{
        \includegraphics[trim={{0cm} {0cm} {0cm} {0cm}}, clip,width=0.85\linewidth]{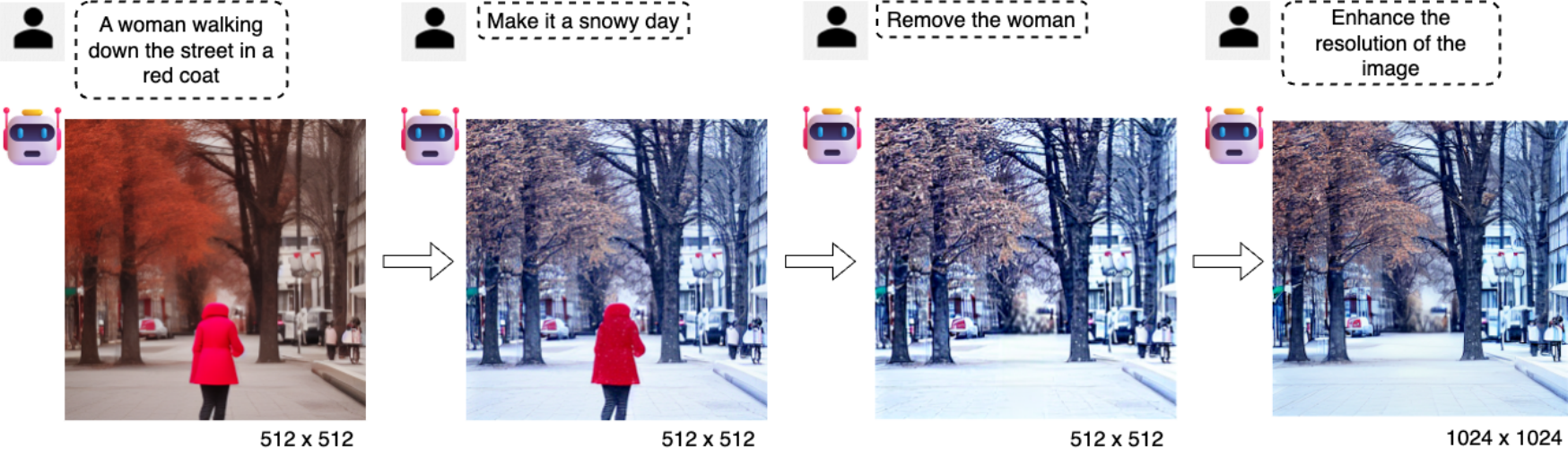}
    }\end{subcaptionbox}
    
    \vspace{-6pt}    
    \begin{subcaptionbox}{}{
        \includegraphics[trim={{0cm} {0cm} {0cm} {0cm}}, clip, width=0.85\linewidth]{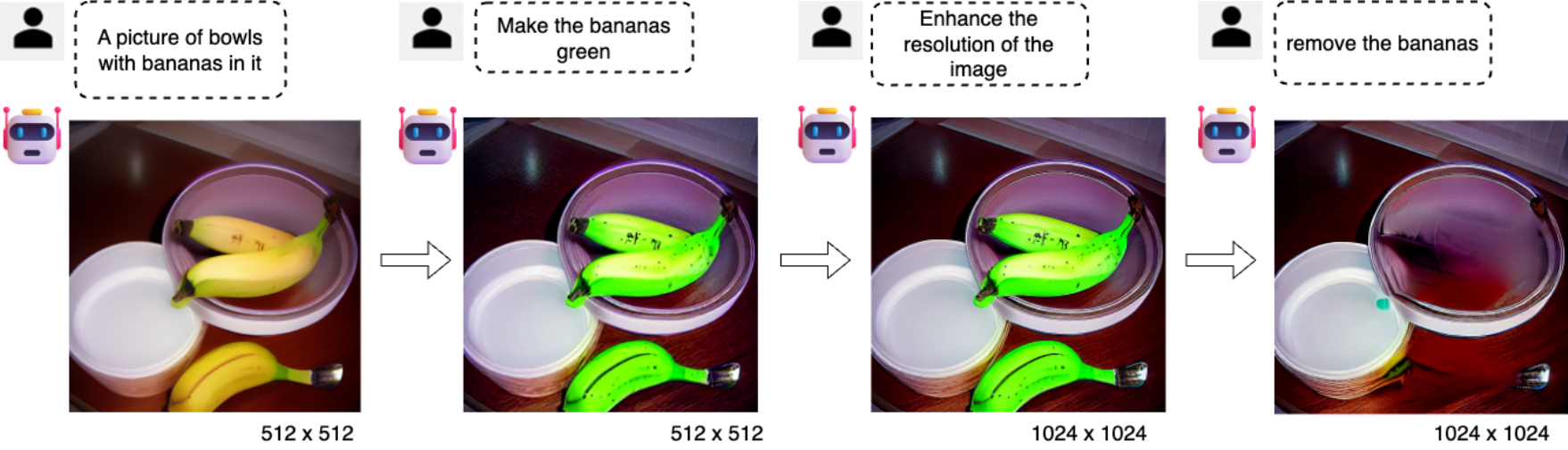}
    }\end{subcaptionbox}  
    \vspace{-10pt}
    \caption{A chatbot showcasing a potential use case of Multi-Task Upcycling. Our approach efficiently upcycles pre-trained text-to-image models, enabling them to perform multiple image generation tasks using a single backbone.}
    \label{fig:chatbot}
\end{figure*}

\begin{abstract}

Text-to-image synthesis has witnessed remarkable advancements in recent years. Many attempts have been made to adopt text-to-image models to support multiple tasks. However, existing approaches typically require resource-intensive re-training or additional parameters to accommodate for the new tasks, which makes the model inefficient for on-device deployment. 
We propose \textit{Multi-Task Upcycling} (MTU), a simple yet effective recipe that extends the capabilities of a pre-trained text-to-image diffusion model to support a variety of image-to-image generation tasks. MTU replaces Feed-Forward Network (FFN) layers in the diffusion model with smaller FFNs, referred to as \emph{experts}, and combines them with a dynamic routing mechanism. To the best of our knowledge, MTU is the first multi-task diffusion modeling approach that seamlessly blends multi-tasking with on-device compatibility, by mitigating the issue of parameter inflation. We show that the performance of MTU is on par with the single-task fine-tuned diffusion models across several tasks including \emph{image editing, super-resolution}, and \emph{inpainting}, while maintaining similar latency and computational load (GFLOPs) as the single-task fine-tuned models. 
\end{abstract}

\vspace{-4mm}
\section{Introduction}

Text-to-image (T2I) generation with diffusion models is rapidly gaining traction across diverse applications, with foundational models such as DALLE2 \citep{ramesh2022}, MidJourney, Stable Diffusion \citep{2022CVPRLDM, Zhang2023ICCV, 2023arXiv230701952P_sdxl, stability_ai_sdxl_turbo, 2024arXiv240213929L_sdxl-lightning}, and Diffusion Transformers (DiT) \citep{2022arXiv221209748P_dit, gao2024lumin-t2x, gao2024lumina-next, xie2024sana, Esser2024ScalingRF} at the forefront. 
Thanks to the open-sourcing efforts, developers have the opportunity to fine-tune them for a variety of creative use-cases. 
The growing demand for generative AI applications has also contributed to the requirement of deploying the {\em state-of-the-art} (SOTA) models on personal and edge devices to address data-privacy and the cost of cloud hosting.

Great effort has been made to optimize these foundation models for edge deployment by making them smaller, faster, and resource-efficient \citep{ZhaoXXJH24, castells2024edgefusion, zhang2024mobile}. 
These optimization strategies include distillation of models to reduce size \citep{xiang2024dkdmdatafreeknowledgedistillation, tang2023lightweightdiffusionmodelsdistillationbased, fang2023structural}, reduction of the frequency of model calls \citep{salimans2022progressive, meng2023distillation, kang2025distilling, zhu2025slimflow, yin2024one, noroozi2024needstepfastsuperresolution}, reduction of on-device memory and latency requirements through quantization \citep{li2023qdiffusion, he2023ptqd, wang2024quest}, and removal of computationally intensive operations~\citep{2023arXiv231116567Z_mobilediffusion}.

As SOTA diffusion models have demonstrated capabilities to support various use cases through fine-tuning, there is also a growing interest to develop a single model to perform multiple image-to-image (I2I) tasks, like image editing \citep{brooks2022instructpix2pix}, super-resolution \citep{Moser_2024}, in/out-painting \citep{Corneanu2024Inpaint, wasserman2024paint}. However, incorporating multiple tasks presents significant challenges. Some approaches adopt universal modeling \citep{diffusionmtl, zhang2024multi, bao2023one} to learn a joint probability distribution that unifies multiple modalities within a common diffusion space. 
Other approaches rely on designing models with specialized components tailored to specific modalities or tasks \citep{xu2022versatile, tang2024any}. 
This allows the diffusion space to vary while partitioning the computational graph based on the task, offering modularity and flexibility. 
However, these methods significantly increase the model size and parameter count, making them computationally inefficient. 
Moreover, these approaches face scalability issue as the computational requirements grow significantly with the increase in the number of tasks and modalities. Thus, a significant gap remains in efficiently adapting diffusion models to multiple tasks, while ensuring they are suitable for on-device deployment.

To bridge the gap, we introduce the concept of {\em Multi-Task Upcycling} (MTU) of T2I diffusion models. MTU transforms single-task T2I models into image-generation generalists, \ie, a single model is capable of handling multiple tasks. 
The concept of {\em upcycling} is widely used in the field of Large Language Models (LLMs) to transform dense pre-trained models into sparse Mixture-of-Experts (MoE) \citep{he2024upcyclinglargelanguagemodels, komatsuzaki2023sparseupcyclingtrainingmixtureofexperts, jiang-etal-2025-improved}, but it has not yet been explored for diffusion models. 
Unlike previous multi-task diffusion models, our approach avoids the need to significantly increase model parameters to support additional new tasks. 
Instead, it only requires retraining a few components with new multi-task data, while ensuring that the model remains both efficient and scalable for multi-task learning. 
MTU is particularly effective in the following two main scenarios: 
\begin{enumerate*}[label=(\roman*), leftmargin=4mm]
    \item extending an on-device T2I diffusion model to support multiple tasks without increasing the computational requirement,
    \item obtaining a multi-task diffusion model with an existing T2I model acting as a strong prior.
\end{enumerate*}
We demonstrate a potential use case of our proposed approach as a \emph{chatbot}, illustrated in Figure \ref{fig:chatbot}, where a single model seamlessly handles multiple tasks such as image editing, inpainting, and super-resolution based on user requests.

Our method is inspired by the {\em empirical} observation that when a T2I model is fine-tuned for a new task, the parameters in the Feed-Forward Network (FFN) layers undergo a significant shift, compared to the other layers of a diffusion model.
Building on this insight and inspired by upcycling literature, we convert a single FFN layer in pre-trained models into a number of smaller FFN \emph{experts}.
These experts are fine-tuned simultaneously on multiple tasks while keeping the rest of the model frozen. A router mechanism dynamically learns to combine the outputs of these individual experts, guided by task-specific embeddings. We evaluate our method on latent diffusion models, such as \emph{SDv1.5} and \emph{SDXL}, across various image-to-image tasks, including image editing, super-resolution, and inpainting. Our results demonstrate that we can develop models that are \emph{iso-FLOP}, i.e., having the same FLOPs as their pre-trained counterparts, while achieving performance comparable to single-task models. Our upcycled SDXL achieves a FID of 3.9 on T2I generation, while maintaining the same computational cost of 1.54 TFLOPs as the pre-trained model.

\section{Related Work}

\noindent
\textbf{Multi-task Diffusion Models:} There are two primary approaches in this area. The first is universal modeling \citep{diffusionmtl, zhang2024multi, bao2023one, chen2024diffusion}, which aims to learn a joint probability distribution unifying multiple modalities within a shared diffusion space. For instance, instead of learning $ p(\text{image}|\text{text}) $ as in single-task diffusion models, it learns $ p(\text{text}, \text{image}) $, the joint probability distribution for text and image. Text-to-image generation can then be performed by marginalizing the joint distribution. However, a key drawback of this approach is its scalability as the entire model needs to be re-trained when new tasks are added. The second approach involves designing models with specialized components tailored to specific modalities or tasks \citep{xu2022versatile, tang2024any}. This enables the diffusion space to vary across tasks, while partitioning the computational graph for modularity and flexibility. Despite these advantages, this method significantly increases model size and parameter count, reducing efficiency. For both approaches, computational requirements scale rapidly with the number of tasks and modalities, resulting in scalability challenges.

\noindent
\textbf{Sparse Upcycling of Pre-trained Models:} Upcycling has been explored in various studies as a method for transforming trained dense models into Mixture-of-Experts (MoE) frameworks \citep{he2024upcyclinglargelanguagemodels, komatsuzaki2023sparseupcyclingtrainingmixtureofexperts, jiang-etal-2025-improved}. The concept of upcycling arose from the challenges of training sparse MoE models from scratch, as such training is highly unstable and sensitive to hyperparameters. Upcycling offers a solution by starting with a pre-trained dense model, which is often readily available online, and transforming it into an MoE model to enhance performance and capacity. At the core of any upcycling method, discussed in \citep{he2024upcyclinglargelanguagemodels, komatsuzaki2023sparseupcyclingtrainingmixtureofexperts, jiang-etal-2025-improved}, lies expert architecture design, initialization techniques, and routing strategies within the MoE layer.

The Feed Forward Networks (FFNs) within pre-trained LLMs, which are two-layer MLPs with hidden dimension $d_{\text{ffn}}$, are replaced by MoE layers. An MoE layer comprises $N$ FFN experts denoted by $\{E_1, E_2, \cdots, E_N\}$ and a router that learns to assign tokens in the input to appropriate experts. Let us denote the hidden dimension of experts as $d_{\text{expert}}$. In sparse upcycling methods, the parameters of these experts are initialized using those of the pre-trained model. Various routing mechanisms have been utilized in different works \citep{pmlr-v139-lewis21a, clark2022unified}, with the Expert Choice router \citep{Zhou2022MixtureofExpertsWE} and standard topK routing \citep{Shazeer2017OutrageouslyLN} being the most commonly used methods for computing the output of the MoE layer.

Many studies have proposed that increasing the number of experts a token is routed to, while simultaneously reducing the dimensions of each expert such that $d_{\text{expert}} < d_{\text{FFN}}$, can be a more efficient approach. This model is referred to as a \emph{fine-grained MoE} model \citep{Krajewski2024ScalingLF} and the ratio $G = d_{\text{FFN}}/d_{\text{expert}}$ is termed as \emph{granularity}. Reducing the dimensions of the experts decreases the FLOPs per expert, which in turn permits an increase in the topK (the number of experts a token is routed to) proportional to the reduction in expert size, all while maintaining the overall FLOPs count. 

We draw inspiration from the concept of \emph{fine-grained MoE} models in the LLM literature and adapt it for multi-tasking in diffusion models. In the following sections, we provide the motivation for focussing on the FFN blocks of diffusion models, followed by a detailed outline of our multi-task upcycling approach.

\section{Preliminaries} \label{sec:preliminaries}

\begin{figure*}[!th]
    \centering
\includegraphics[trim={{1cm} {2.8cm} {3cm} {3cm}}, clip,width=0.97\linewidth]{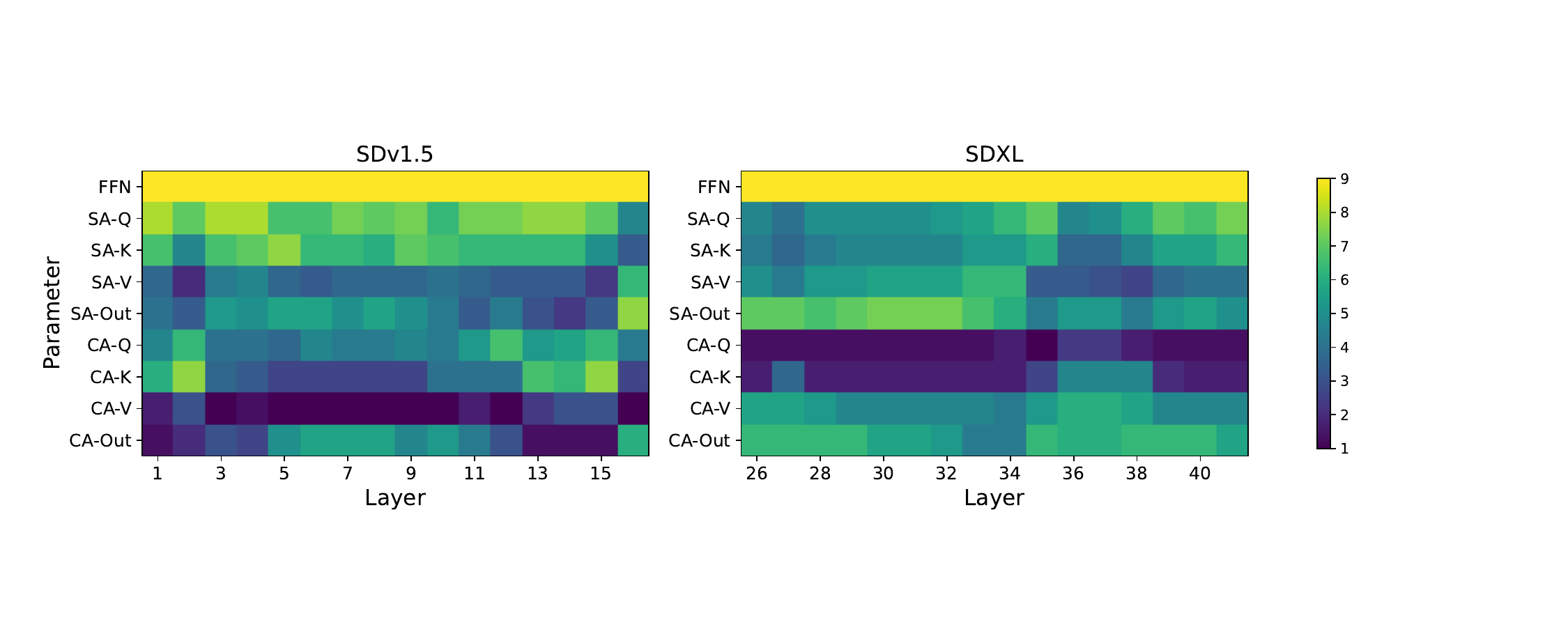}
\vspace{-4mm}
\caption{We analyze the deviation between fine-tuned weights $\theta_{f}^{\tau}$ and pre-trained initialization $\theta_{p}$ across different layers in the LDM (i.e., $\deW = || \theta_{f}^{\tau} - \theta_{p} ||$) and rank them accordingly. 
We present the average rank of these deviations across all tasks. The x-axis represents layer depth, while the y-axis indicates the component type. 
FFN layers show the highest deviation, suggesting they specialize in adapting to downstream tasks.}
    \label{fig:parameter-diff}
\end{figure*}

\textbf{Latent Diffusion Models:} Latent Diffusion Models (LDMs) \citep{2022CVPRLDM, Zhang2023ICCV} are based on Diffusion Models (DMs) \citep{ho2020denoising, song2021denoising}, that learn to reverse a forward Markov process in which noise is incrementally added to input images over multiple time steps $t \in [0, T]$. We denote an RGB image by $\mathbf{x}_0 \in \Real^{3 \times H \times W}$, where $H$ and $W$ correspond to the height and width of the image. An encoder $\mathcal{E}$ transforms the input image $\mathbf{x}_0$ into a latent representation $\mathbf{z}_0 \in \Real^{c \times h \times w}$, where $h$ and $w$ represent the height and width of the downscaled encoded image, and $c$ indicates the number of channels in latent space. During training, a noisy latent $ \mathbf{z}_t$ at time $t$ is obtained from a real image's latent $\mathbf{z}_0$ by $\mathbf{z}_t = \sqrt{a_t} \mathbf{z}_0 + \sqrt{1-a_t} \mathbf{\epsilon}$, where $\mathbf{\epsilon} \sim \mathcal{N}(\mathbf{0}, \mathbf{I})$ and $a_t$ is a parameter that gradually decays over time. A denoiser $f_\theta(.)$ is then trained to predict the noise added to $\mathbf{z}_t$ conditioned on the input text embedding $\mathbf{c}_T$. This enables the reconstruction of $\mathbf{z}_0$ by subtracting the predicted noise from $\mathbf{z}_t$. To achieve this, the denoiser is trained to predict the noise by stochastically minimizing the objective: $\mathcal{L}(\mathbf{z}, \mathbf{c}_T)=\mathbb{E}_{\epsilon, \mathbf{x}, \mathbf{c}_T, t}\left[\left\|\epsilon-f_\theta\left(\mathbf{z}_t, \mathbf{c}_T, t\right)\right\|\right]$.
A decoder $\mathcal{D}$, then maps the denoised $\hat{\mathbf{z}_0}$ back to the pixel space. During inference, given a text prompt $\mathbf{c}_T$, a noisy latent embedding $\mathbf{z}_T$ is sampled and iteratively denoised over $T$ steps to produce $\hat{\mathbf{z}_0}$, which is decoded into the final image. Typically, the encoder and decoder are derived from a pre-trained autoencoder that remains frozen during training.

\textbf{Fine-tuning LDMs for Image-to-Image Generation Tasks:} The objective of image-to-image (I2I) generation tasks is to transform an input image $\mathbf{c}_I$ into a target image $\mathbf{c}_{\text{target}}$ based on an edit prompt $\mathbf{c}_T$. In I2I literature \citep{brooks2023instructpix2pix}, $\mathbf{c}_T$ and $\mathbf{c}_I$ are commonly referred to as the text and image \textit{conditions}, respectively. The target image and input image are encoded by an encoder $\mathcal{E}$ to obtain the latent representations $\mathbf{z}_{\text{target}}$ and $\mathbf{z}_c$, respectively. To train a diffusion model, noise is added to $\mathbf{z}_{\text{target}}$ to obtain $\mathbf{z}_t$ by $\sqrt{a_t} \mathbf{z}_{\text{target}} + \sqrt{1-a_t} \mathbf{\epsilon}$, where $\mathbf{\epsilon} \sim \mathcal{N}(\mathbf{0}, \mathbf{I})$ and $a_t$ is a parameter that is scheduled similar to T2I diffusion models. A denoiser $f_{\theta}$ is then trained to predict the noise added to a noisy input latent $\mathbf{z}_{\text{target}}$, given the image condition $\mathbf{z}_c$ and the text instruction $\mathbf{c}_T$. To achieve this, the image condition $\mathbf{z}_c$ is concatenated with the noisy latents $\mathbf{z}_t$, and the resulting tensor is provided as input to the denoiser.  To accommodate the additional channels introduced by the image conditioning, the first convolutional layer is modified to include extra input channels, while the rest of the architecture remains unchanged. The training process involves minimizing the following latent diffusion objective: $\mathcal{L}(\mathbf{z}, \mathbf{c}_T, \mathbf{z}_c)=\mathbb{E}_{\epsilon, \mathbf{x}, \mathbf{c}_T, \mathbf{z}_c, t}\left[\left\|\epsilon-f_\theta\left(\mathbf{z}_t, \mathbf{c}_T, \mathbf{z}_c, t\right)\right\|\right]$. Pre-trained text-to-image models, such as Stable Diffusion, are commonly used as initialization to leverage their extensive text-to-image generation capabilities.

\section{Motivation}
\label{sec:motivation}

\begin{table}[t]
    \centering
    \small
    \resizebox{0.99\linewidth}{!}{\begin{tabular}{c|c|c|c}
    & Image Editing (IE) & Super Resolution (SR) & Inpainting (IP) \\
    \midrule
    & I-T Dir Sim $\uparrow$ & LPIPS $\downarrow$ & I-I Dir Sim $\uparrow$ \\
    \midrule
    SA & 17.6 & 25.0 & 42.9 \\
    CA & 16.9 & 27.5 & 40.1\\
    FFNs & \textbf{17.8} & \textbf{23.7} & \textbf{46.7}\\
    \end{tabular}}
    \caption{Quantitative comparison of fine-tuning different components (SA, CA, and FFNs) of the diffusion model over image-to-image generation tasks. Fine-tuning FFNs leads to better performance on image-to-image tasks.}
    \label{tab:ft_motivation}
\end{table}

\begin{figure*}[t]
    
    \centering
    \begin{overpic}[width=\textwidth]{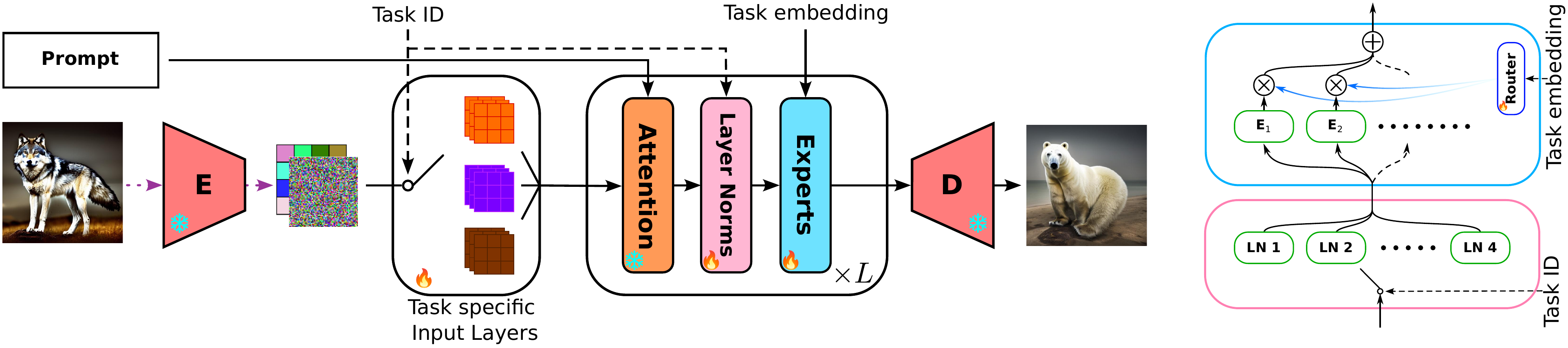}
        \put(88,30){\tiny $\mathcal{N}(0,1)$}
        \put(0,102){\textbf{a)}}
        \put(370,102){\textbf{b)}}
    \end{overpic}
    \caption{(a) Overview: We introduce Multi-task Upcycling (MTU), a method for transforming a pre-trained text-to-image model to support multiple tasks. (b) In MTU, we replace the FFN layer in the pre-trained model with a set of smaller experts, which are dynamically combined using a router mechanism.}\label{fig:method}
\end{figure*}

Previous studies~\citep{loshchilov2017sgdr, overcoming2017james} in the area of transfer learning and domain adaptation have shown that parameters undergoing significant changes during fine-tuning are more relevant to the specific task, while those with minimal changes are either already well-aligned with the task due to pre-training or less critical for fine-tuning. 
Inspired from these findings, we identify a subset of parameters within the diffusion models that show the highest deviations from the pre-trained initialization. In this experiment, we consider two Latent Diffusion Models, namely SDv1.5 and SDXL. We fine-tune them on three tasks: (i) image editing, (ii) super-resolution, and (iii) inpainting. See the appendix for further details on the dataset and training procedure in Section \ref{sec:app-training}.

Let $\theta_{p}$ be the weights of a pre-trained diffusion model, and $\theta_{f}^{\tau}$ the fine-tuned weights for a specific task $\tau$. The distance between the fine-tuned weights and the initialization is computed as $\deW = || \theta_{f}^{\tau} - \theta_{p} ||$, where $|| \cdot ||$ is the Frobenius norm.
For each task, we define $\deW$ for all LDM components: Self-Attention ($\deW_{\text{SA}}$), Cross-Attention ($\deW_{\text{CA}}$), and Feed-Forward Network ($\deW_{\text{FFN}}$) blocks. Within the SA and CA blocks, we further consider the Q, K, V, and output layer matrices. 
We rank $\deW_{\text{SA}}$, $\deW_{\text{CA}}$, and $\deW_{\text{FFN}}$ for each layer across all tasks and then compute the average rank across I2I tasks. Figure~\ref{fig:parameter-diff} illustrates layer-wise average ranks for I2I tasks.
Higher ranks signify greater deviation from the pre-trained initialization. Our experiments show that FFN layers deviate the most from the pre-trained weights to adapt to downstream I2I tasks. This is also demonstrated in Figure \ref{fig:sd-distances} in the appendix, which presents measured deviation values.

To further validate this observation, we fine-tune each component of the attention block separately and compare their performance. Details on the metrics reported are in Section \ref{sec:experiment-details}. As shown in Table~\ref{tab:ft_motivation}, fine-tuning the FFN layers in diffusion models consistently yields better results compared to tuning other components. In other words, FFNs specialize in solving a downstream task, while the other parameters in the attention block learn more general features. In this paper, we build on these findings, and propose an approach for Multi-Task Upcycling for Diffusion Models.




\section{Methodology: Multi-task Upcycling for Diffusion Models} 
\label{sec:method}


In MTU, we segment the FFN component into experts such that their weighted combination solves specific tasks without the need of extra parameters. 
MTU comprises four key steps: 
\begin{enumerate*}[label=(\roman*), leftmargin=4mm]
\item split the pre-trained model's FFNs into smaller FFN experts, 
\item design a router to dynamically combine the outputs of these experts, 
\item define task-specific input processing layers, and 
\item design the loss function to train the FFN experts and the router. 
\end{enumerate*}
Note that only the FFN experts, the router, and the task-specific input processing layers are trainable parameters. An overview of our method is presented in Figure~\ref{fig:method}.

\noindent
\textbf{Expert architecture:} We replace each FFN block in the pre-trained denoiser architecture with $N$ experts $\{\expert_1^l, \cdots, \expert_N^l\}$, each with dimension $d_{\text{expert}}^l$, where $l$ is the layer index. To ensure that the upcycled model maintains a parameter count similar to the pre-trained model, we constrain $d_{\text{expert}}^l = d_{\text{FFN}}^l/N$, where $d_{\text{FFN}}^l$ denotes the dimension of $l$-th FFN block in the pre-trained model. 

Upcycling methods typically initialize the parameters of the experts from the pre-trained model. However, in our case, since the size of the FFN in the pre-trained model differs from the upcycled model, copying the dense MLP weights to the upcycled model structure becomes non-trivial. To address this, we segment the dense layer into $N$ shards along the appropriate dimension and copy each shard into the corresponding expert. We denote the set of parameters in all the experts within the entire model as $\theta_{\expert}$.

\noindent
\textbf{Router:} Let $\mathcal{T}$ denote the set of tasks. For each task $\tau \in \mathcal{T}$, we define a learnable task embedding $e_{\tau} \in \Real^{d_\text{task}}$. 
We add a router to a layer $l$ as: $g(\cdot \ ; \ \theta_r^l): \{ \{e_{\tau} \}_\mathcal{T} \rightarrow \Real^N$\}, where each expert is assigned a weight based on $e_{\tau}$. The output of the FFN block for an input $x_\tau$ corresponding to task $\tau$ is computed as follows:
\begin{gather}
\small
w_i^l = \text{softmax}(g(e_{\tau}; \theta_r^l))_i \nonumber \\
E_{\text{FFN}}^l(x_\tau) = \sum_{i=1}^N w_i^l \times E_i^l(x_\tau),
\label{eq:expert_combination}
\end{gather}

where $w_i^l$ represents the weight assigned to the $i$-th expert, and $E_i(x_\tau)^l$ is the output of the $i$-th expert for input $x_\tau$ in layer $l$ of the model. A key advantage of this formulation is that the task-specific weights can be pre-calculated with minimal computational overhead using the task identifier.

\noindent
\textbf{Task-specific Layer Norms:} Our empirical observation suggests that incorporating a task-specific layer normalization step before each expert significantly improves model performance. The arrangement of these layer normalization steps is illustrated in Figure~\ref{fig:method}(b). Our experiments show that FFN layers exhibit significantly different distributions across tasks, and thus the task-specific layer normalization facilitates the learning of these distinct distributions in the upcycled model. We denote the set of all task-specific convolution layers as $\Psi_{L}$.

\noindent
\textbf{Task-specific input layers:} As outlined in Section~\ref{sec:preliminaries}, single-task image-to-image models share the same architecture, differing only in the input convolution layer, which accommodates additional channels from image conditioning. In our upcycled architecture, we introduce separate task-specific input convolution layers, denoted by $\psi_{\tau}(\cdot)$, to handle the varying conditioning distributions for different tasks. We denote all the set of task-specific convolution layers as $\convW$.

\noindent
\textbf{Multi-task Loss:} 
For a task $\tau$, let $z^{\tau}_0$ represent the encoded images, $z_t^\tau$ the noisy latents at time step $t$. We feed $z_t^\tau$ and the text prompt $c_T^\tau$ to the denoiser $f$, and task it to predict the noise $\epsilon$. Note that in the case of image conditioning, $z^\tau_t$ is the concatenation of the noisy latents and the encoded context image $c_I^\tau$ along the channel dimension.

To reduce the computational burden, we pre-compute the expert weights using the task identifiers $e_{\tau}$. We use these weights to combine the expert outputs as shown in Equation~\ref{eq:expert_combination}. While the shared parameters of the model are kept frozen, the task-specific layer norms $\Psi_L$ and input layers $\Psi_C$, experts $\theta_E$ and routers $\theta_R$ are trained to optimize a multi-task objective with the loss $L$ defined as: 

\begin{equation}
\small
    L=\sum_{\tau \in \mathcal{T}}\mathbb{E}_{\epsilon, \mathbf{x}^\tau, \mathbf{c}_T^\tau, \mathbf{c}_I^\tau, t} \left\|\epsilon-f_\theta\left(\convW(\mathbf{z}_t^\tau), \mathbf{c}_T^\tau, \mathcal{E}(\mathbf{c}_I^\tau), t\right)\right\| 
\end{equation}


\section{Experimental Settings}
\label{sec:experiment-details}

\begin{table*}[t]
    \centering
    \small
    \resizebox{\linewidth}{!}{
    \begin{tabular}{c|c|c|cc|c|c|c|c}
    \toprule
         & Multi-task & Model &  TFLOPs & Parameters &  Text-to-Image (T2I) & Image Editing (IE) & Super Resolution (SR) & Inpainting (IP)\\
          \midrule
         & & & & & FID $\downarrow$ & I-T Direction Similarity $\uparrow$ & LPIPS $\downarrow$ & I-I Directional Similarity $\uparrow$ \\
        \midrule
         \multirow{5}{*}{\rotatebox{90}{SD v1.5}} & \multirow{4}{*}{\textcolor{red}{$\times$}} & T2I \citep{2022CVPRLDM} & \multirow{4}{*}{0.67} & \multirow{4}{*}{860M} & 12.9 & -- & -- & -- \\
        & & IE \citep{brooks2022instructpix2pix} & & & -- & 15.4 & --& -- \\
        & & SR \citep{2022CVPRLDM} & & & -- & -- & 38.0 & --\\       
        & & IP \citep{yildirim2023instinpaint} &  & & -- & -- & -- &  \textbf{46.5} \\
        \cmidrule{2-9}
        & \multirow{2}{*}{\textcolor{ForestGreen}{$\checkmark$}} & VD \citep{xu2022versatile} & 0.87  & 1.1B & 10.1 & 14.2 & -- & --\\
        & & Unidiffuser \citep{bao2023one} & 0.83  & 952M & 7.4 & -- & -- & -- \\
        \cmidrule{2-9}
        & \textcolor{ForestGreen}{$\checkmark$} & MTU (Ours) & 0.68 & 869M & \textbf{7.2}  & \textbf{17.2} & \textbf{24.8} &  44.0  \\
        \midrule
        \multirow{5}{*}{\rotatebox{90}{SDXL}} & \multirow{4}{*}{\textcolor{red}{$\times$}} & T2I \citep{2023arXiv230701952P_sdxl}& \multirow{4}{*}{1.53} & \multirow{4}{*}{2.6B} & 4.1 & -- & -- & -- \\
        &  & IE \citep{brooks2022instructpix2pix} &  & & -- & 17.3 & --& -- \\
        &  & SR & & & -- & -- & 26.9  & --\\
        &  & IP & & & -- & -- & -- & 43.2\\
        \cmidrule{2-9}
        & \textcolor{ForestGreen}{$\checkmark$} & MTU (Ours) &  1.54& 2.6B & \textbf{3.9} & \textbf{20.1} & \textbf{26.5} & \textbf{44.2}  \\
    \bottomrule
    \end{tabular}}
    \caption{Quantitative comparison of Multi-task Upcycling (MTU) against single-task and multi-task baselines. We consider $N=4$ for SDv1.5 and $N=1$ for SDXL. MTU consistently surpasses baselines while preserving computational efficiency.}
    \label{tab:main-results}
\end{table*}

\noindent
\textbf{Model Architectures:} We evaluate our method on two Stable Diffusion-based models—SDv1.5 \citep{2022CVPRLDM} and SDXL \cite{2023arXiv230701952P_sdxl} consisting of 860M and 2.6B parameters in the denoiser component respectively. In both models, the denoiser is a UNet composed of transformer blocks with Self-Attention (SA), Cross-Attention (CA), Feed-Forward Networks (FFNs), and residual blocks. For multi-task upcycling, we consider all transformer blocks in SDv1.5 (16 blocks) and SDXL (70 blocks).

Each router network $ g(\cdot~|~\theta_r^l) $ is implemented as a two-layer MLP with ReLU activation. As shown in Equation \ref{eq:expert_combination}, we then apply a softmax function over the router's predictions to obtain the weights assigned to each expert.

\noindent
\textbf{Downstream Image Synthesis Tasks:} We consider four tasks in our study, including Text-to-Image (T2I) generation, Image Editing (IE), Super Resolution (SR), and Image Inpainting (IP). 
Since the MTU model is initialized with a pre-trained T2I model, we include T2I as one of the tasks to ensure the multi-task model maintains its text-to-image generation capability.

\noindent
\textbf{Datasets:} Since each task requires different data configurations, we use the following datasets to train the MTU model. 
\vspace{-4mm}
\begin{itemize}[leftmargin=*]
\setlength{\itemsep}{1pt}
\setlength{\parskip}{0pt}
    \item T2I: We use the COCO Captions dataset~\citep{lin2015microsoftcococommonobjects}, a large collection of image-text pairs.
    
    \item Image Editing: We use the dataset introduced in \citep{brooks2022instructpix2pix}, which provides input and target images along with corresponding editing instructions.

    \item Super Resolution: We use the Real-ESRGAN dataset \citep{wang2021real}, which consists of high-resolution images. We generate corresponding low-resolution images by applying degradations and downscaling them by half. 

    \item Image Inpainting: We use the dataset from \citep{yildirim2023instinpaint}, which provides a multi-modal inpainting dataset designed for object removal based on text prompts.
\end{itemize}
\vspace{-4mm}
More details can be found in Section \ref{sec:app-datasets} in the Appendix.




\noindent
\textbf{Training Details:} We freeze all other layers and train only the FFN experts, routers, and task-specific layers, as described in Section \ref{sec:method}. This results in training 158M parameters for SDv1.5 and 1.5B parameters for SDXL. Both models are trained on 8× A100 GPUs for 100 epochs, with a batch size of 16 per GPU and image resolution of 512 $\times$ 512. SDXL is optimized using AdamW with a learning rate of 5e-5, while SDv1.5 is trained using Adam with a learning rate of 1e-4. During sampling, we perform denoising for 20 and 50 iterations for multi-task SDv1.5 and SDXL models respectively. More details are presented in Section \ref{sec:app-training} in the appendix.



\noindent
\textbf{Metrics:} We used the following metrics that are commonly considered for evaluating models train for specific tasks. 
\vspace{-4mm}
\begin{itemize}[leftmargin=*]
\setlength{\itemsep}{1pt}
\setlength{\parskip}{0pt}
    \item T2I: We report Frechet Inception Distance (FID) on the test set of the COCO captions dataset. Lower values indicated better images.
    
    \item SR: We measure Learned Perceptual Image Patch Similarity (LPIPS) \citep{zhang2018perceptual} between generated and ground truth images, where lower values indicate more similarity. 
    
    \item Image Editing: We report Image-Text (I-T) Directional Similarity \citep{brooks2022instructpix2pix}, which quantifies how well the change in text captions aligns with corresponding edits. Let $I_{\text{input}}$ and $I_{\text{edited}}$ represent CLIP-extracted features of input and edited images respectively \citep{radford2021learningtransferablevisualmodels}. Similarly, let $T_{\text{edited}}$ and $T_{\text{input}}$ denote the CLIP-extracted text features for the input and edited descriptions. I-T Directional Similarity is defined as $S(T_{\text{edited}} - T_{\text{input}}, I_{\text{edited}} -  I_{\text{input}})$, where $S$ is the cosine similarity.
    
    \item Inpainting: We report Image-Image (I-I) Directional Similarity, which measures alignment with the ground truth. Given CLIP features corresponding to ground truth image $I_{\text{gt}}$, I-I Directional Similarity is defined as $S(I_{\text{gt}} - I_{\text{input}}, I_{\text{edited}} -  I_{\text{input}})$. Higher values indicate better similarity to the ground truth.

\end{itemize}


\section{Results}


 \begin{figure*}[] 
 \captionsetup[subfigure]{labelformat=empty}
    \centering
    \begin{subcaptionbox}{}{
        \includegraphics[width=0.45\textwidth]{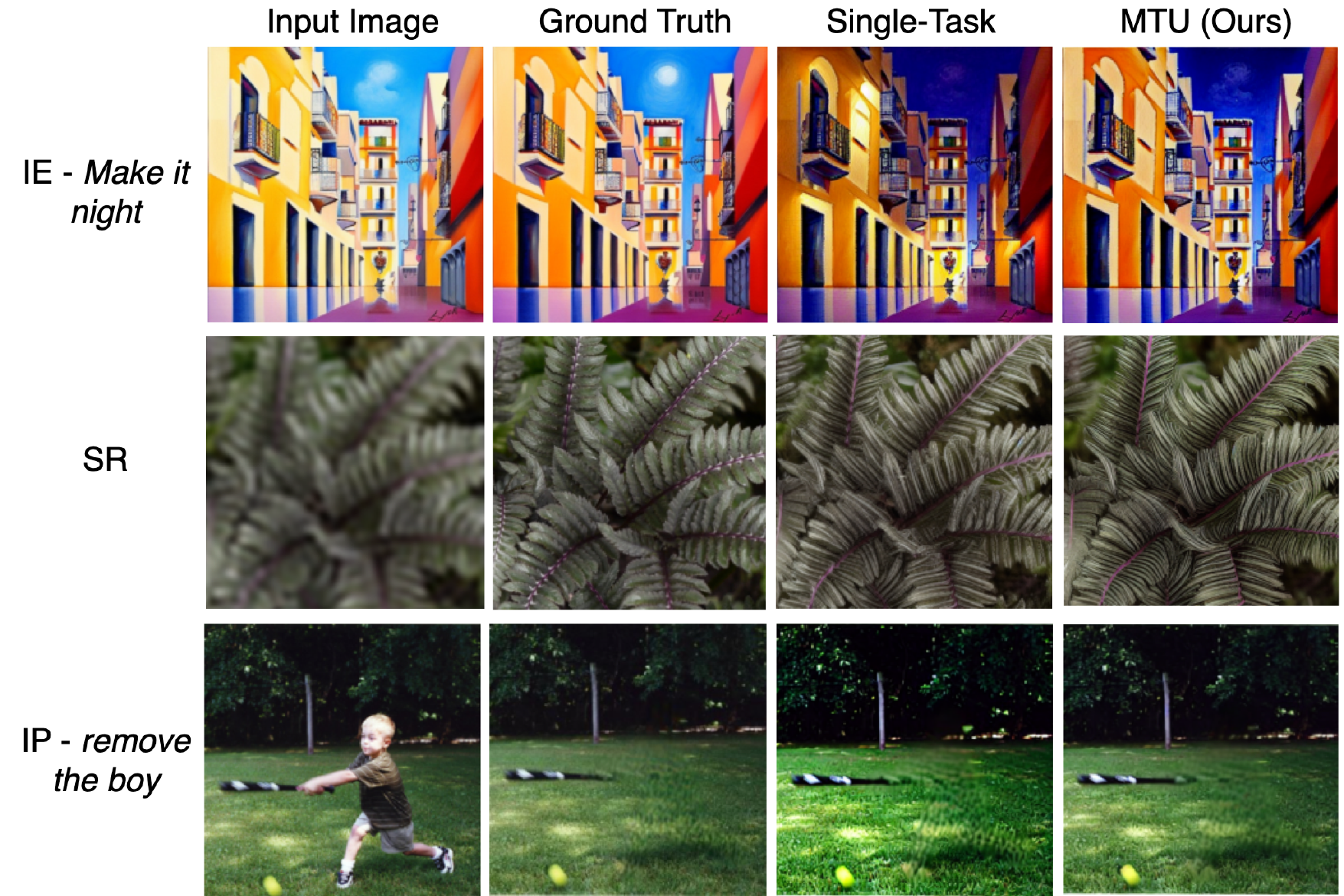}
    }\end{subcaptionbox}
    \begin{subcaptionbox}{}{
        \includegraphics[width=0.45\textwidth]{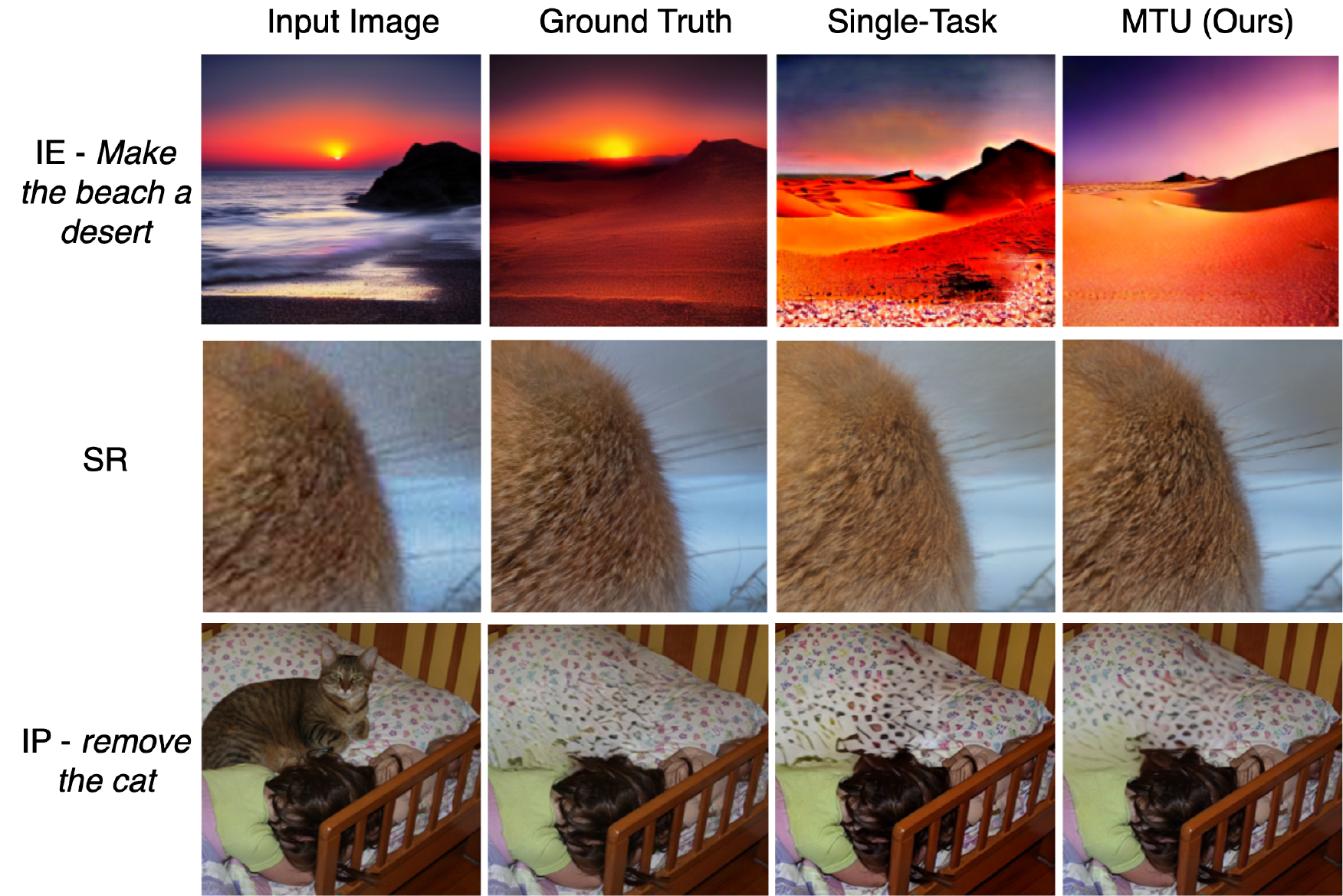}
    }\end{subcaptionbox}
    \vspace{-4mm}
    \caption{Qualitative comparison of MTU based on SDv1.5 (left) and SDXL (right) with corresponding single-task baselines for Image Editing (IE) \citep{brooks2022instructpix2pix}, Super Resolution (SR) \citep{2022CVPRLDM}, and Inpainting (IP). \citep{yildirim2023instinpaint}}
    \label{fig:qualitative-results}
\end{figure*}

\begin{figure}
    \centering
    \includegraphics[trim={{0cm} {0.7cm} {3cm} {1cm}}, clip,width=\linewidth]{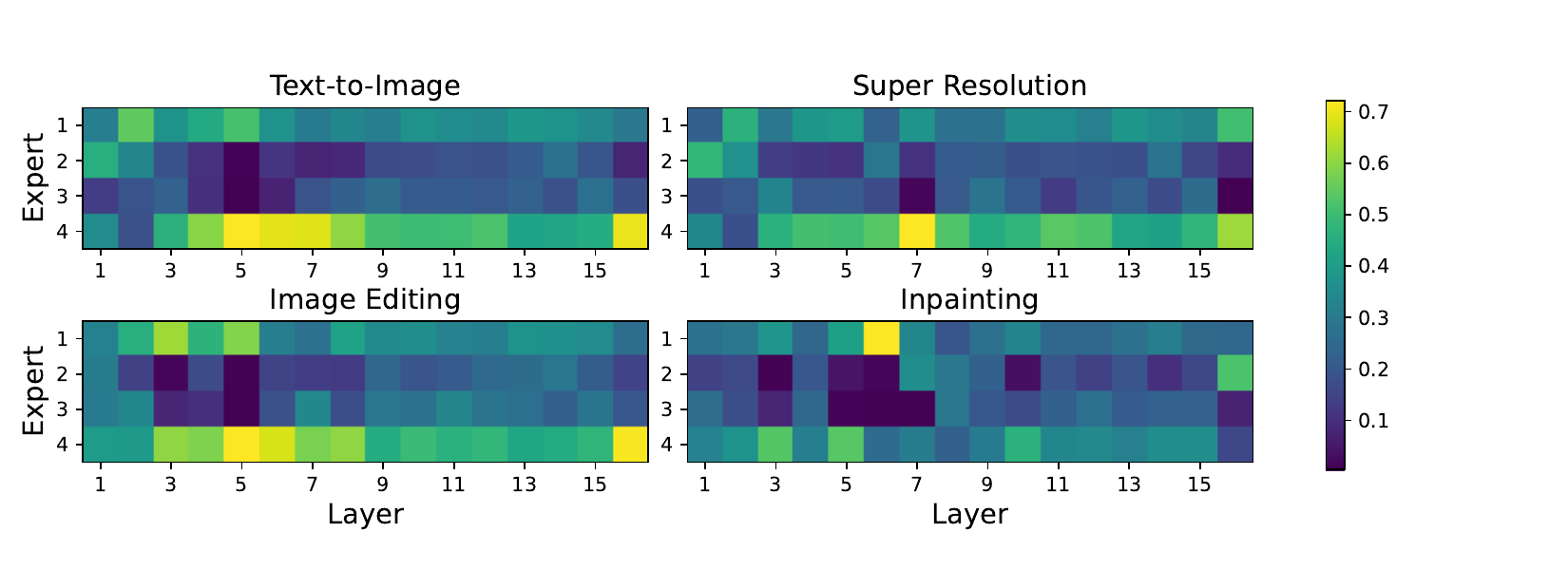}
    \vspace{-6mm}
    \caption{Analysis for expert selection by the router. We show the expert weight distribution assigned by the router for SDv1.5 with four experts (N = 4).
    }
    \vspace{-5mm}
    \label{fig:expert-selection}
\end{figure}

\textbf{Comparison with single-task and multi-task models} \\We compare MTU against both single-task and multi-task baselines. For single-task baselines, we prioritize open-source models based on SDv1.5 or SDXL whenever available. If no open-source models are available, we fine-tune the pre-trained T2I model on the I2I task following the methodology outlined by \citep{brooks2022instructpix2pix} in Section \ref{sec:preliminaries}. 
Specifically, we consider the image editing SDv1.5 and SDXL models fine-tuned by \citep{brooks2022instructpix2pix}. For Super-Resolution (SR) and Inpainting, we could only compare SDv1.5 based models from \citep{2022CVPRLDM} and \citep{yildirim2023instinpaint}, respectively. Since no equivalent open-source versions are available for SDXL, we fine-tuned a model locally for both SR and Inpainting.


Additionally, we use the Versatile Diffusion (VD) \citep{xu2022versatile} and UniDiffuser \citep{bao2023one} models as multi-task model baselines, as they are, to the best of our knowledge, the only models comparable to the MTU model. 
Note that these multi-task models were trained under entirely different settings and do not cover all the tasks considered in this paper. Therefore, for a fair comparison, we evaluate UniDiffuser solely on the T2I task and Versatile Diffusion on the T2I and Image Editing tasks.

Table \ref{tab:main-results} shows that our method consistently outperforms both single-task and multi-task baselines. Specifically, MTU based on SDv1.5 achieves scores of 17.2 in Image Editing and 24.8 in Super-Resolution (SR), while single-task baselines lag by 3 and 14 points, respectively. Additionally, MTU based on SDv1.5 outperforms existing multi-task baselines, further demonstrating its effectiveness. We also report the compute requirements for each model in GFLOPs, showing that our approach maintains computational efficiency comparable to single-task models. Qualitative results are presented in Figure \ref{fig:qualitative-results} and Section \ref{sec:app-pictures} in the appendix.

\begin{table}[]
    \small
    \centering
    \resizebox{0.99\linewidth}{!}{
    \begin{tabular}{cc|c|c|c}
    \toprule
     & & IE & SR & IP \\
     \midrule
        & & I-T Dir Sim $\uparrow$ & LPIPS $\downarrow$ & I-I Dir Sim $\uparrow$ \\
        \midrule
        \multirow{3}{*}{\rotatebox{0}{SDv1.5}}
        & LORA & 8.3 & 43.3 & 31.4\\
        & IA3 & 11.2 & 39.3 & 34.2  \\
        & Full - FT &15.4 & 38.0 & 46.5 \\
        \midrule
        &  MTU ($N=4$) &  \textbf{17.2} & \textbf{24.8} & \textbf{44.0} \\
        \midrule
        \multirow{3}{*}{\rotatebox{0}{SDXL}}
        & LORA & 13.1 & 31.8 & 28.9\\
        & IA3 & 8.4 & 51.0 &  30.6 \\
        & Full - FT & 17.3 & 26.9 & 43.2 \\
        \midrule
        &  MTU ($N=1$) & \textbf{20.1} & \textbf{26.5} & \textbf{44.2}\\
    \bottomrule
    \end{tabular}
    }
    \vspace{-2mm}
    \caption{Comparison with PEFT methods. MTU significantly outperforms PEFT methods applied on FFN layers.}
    \vspace{-4mm}
    \label{tab:peft-results}
\end{table}

\noindent
\textbf{Comparison with PEFT methods} \\
Parameter-efficient fine-tuning (PEFT) methods are widely used to adapt pre-trained models to different tasks. Methods like LoRA \citep{hu2022lora} and IA3 \citep{liu2022fewshot} are lightweight, requiring only a small fraction of parameters to be fine-tuned within task-specific adapters. 
Since our analysis in Section \ref{sec:motivation} highlights that FFNs play a crucial role in enabling support for image-to-image tasks, we compare MTU with LoRA and IA3 for tuning the FFN layers. 
Note that applying IA3 only to FFN blocks did not yield good results, so we apply IA3 to the entire block including self-attention, cross-attention and FFN layers. 
By comparing our approach with these PEFT methods, we evaluate how effectively our method performs in comparison to efficient fine-tuning techniques that primarily target FFN layers. Table \ref{tab:peft-results} compares MTU with PEFT methods such as LoRA and IA3. For SDXL, our method outperforms IA3 by approximately 6 absolute points in image editing and 15 absolute points in super-resolution (SR). Overall, MTU consistently outperforms LoRA and IA3 across all tasks, demonstrating better multi-task support compared to task-specific adapters.

\noindent
\textbf{Analysing router assignment for tasks} \\
We analyze how the router distributes weights across the set of experts for each task to determine which experts specialize in specific tasks. Figure \ref{fig:expert-selection} illustrates the expert weight distribution assigned by the router for SDv1.5 with $N=4$.  Our findings show that Text-to-Image (T2I) generation and Image Editing share three experts: $E_4^5$, $E_4^6$, and $E_4^{16}$. Meanwhile, $E_4^7$ specializes in Super-Resolution (SR), and $E_1^6$ is assigned greater importance for Image Inpainting. The weight distribution across experts is similar for T2I (trained on COCO) and Image Editing, as both tasks require strong prompt-following capabilities. In contrast, Inpainting primarily relies on object removal prompts, while SR operates without any textual conditioning, necessitating the use of different experts for these tasks.

\begin{table}[]
    \centering
    \resizebox{\linewidth}{!}{
    \begin{tabular}{ccc|c|c|c|c}
    \toprule
     & \# experts $N$ & top-$k$ & T2I & IE & SR & IP \\
     \midrule
        & & & FID $\downarrow$ & I-T Dir Sim $\uparrow$ & LPIPS $\downarrow$ & I-I Dir Sim $\uparrow$ \\
        \midrule
        \multirow{5}{*}{\rotatebox{90}{SDv1.5}}
        & 1 & -- & \phantom{0}7.3 & 17.6 & 25.3 & 44.0 \\
        & 2 & -- & \phantom{0}8.4 & \phantom{0}9.0 & 30.9 & 43.2 \\
        & 4 & -- & \phantom{0}\textbf{7.2} & \textbf{17.2} & \textbf{24.8} & \textbf{44.0}\\
        & 8 & -- & \phantom{0}8.0 & 10.2 & 27.8 & 38.7 \\
        & 8 & 4 & \phantom{0}7.9 & \phantom{0}8.9 & 25.8 & 39.3 \\
        & 16 & -- & \phantom{0}8.5 & \phantom{0}9.6 & 27.2 & 36.5 \\
        & 16 & 4 & \phantom{0}8.2 & \phantom{0}8.6 & 26.5 & 38.6  \\
        \midrule
        \multirow{5}{*}{\rotatebox{90}{SDXL}}
        & 1 & -- & \phantom{0}\textbf{3.9} & \textbf{20.1} & \textbf{26.5} & \textbf{44.2} \\
        & 2 & -- & 10.5 & 10.4 & 30.4 & 39.9 \\
        & 4 & -- & 12.3 & 11.8 & 30.5 &  38.6 \\
        & 8 & -- & 12.8 & 12.4 & 31.6 & 34.5\\
        & 8 & 4 & 13.6 & 12.8 & 31.5 & 32.7 \\
        & 16 & -- & 17.1 & 11.8 & 32.8 & 31.7 \\
        & 16 & 4 & 18.2 & 10.9 & 33.2 & 31.9\\
    \bottomrule
    \end{tabular}}
    \caption{Performance of MTU on SDv1.5 and SDXL with varying number of experts  $N$.}
    \label{tab:ablation-results}
\end{table}

\noindent
\textbf{MTU models with varying number of experts} \\
We conduct an ablation study on the number of experts in the MTU model. Since our objective is to maintain a parameter count similar to the pre-trained model, the hidden dimension of each expert decreases as we increase the number of experts. Table \ref{tab:ablation-results} presents the performance of our method with varying number of experts. Our findings show that for both SDv1.5 and SDXL, increasing the experts to a higher number (consequently decreasing expert dimensions) significantly degrades performance. For SDv1.5, the optimal number of experts is N = 4, while for SDXL, dividing the model into experts is suboptimal, as tuning the task-specific layer norms and FFN blocks alone provide sufficient multi-task support. Given that MTU for SDXL works best with a single expert, we conducted an additional experiment where FFNs were frozen, and only the layer norms preceding them were fine-tuned. However, this model failed to converge, implying that FFNs play a key role in learning I2I tasks.

It is worth noting that an alternative approach is to increase the number of experts beyond the total parameter count of the pre-trained model while selecting only the top-$k$ experts per task, ensuring that the number of active parameters—i.e., the parameters used for a single task—remains the same. To evaluate this approach, we increase the number of experts to $8$ and $16$ while selecting the top $4$ experts such that the total number of parameters are same as the original FFN layer. As shown in Table~\ref{tab:ablation-results}, we observe that this approach for increasing the experts with limited parameter count does not perform on par with simply combining all of the experts.

\section{Discussions}


In this section, we provide recommendations for upcycling a pre-trained text-to-image model into a multi-task image generation model. Until now, PEFT methods are often the first choice for practitioners seeking to enable multi-task support due to their computational efficiency. However, our experiments with task-specific PEFT methods show that while they are lightweight, our MTU approach consistently outperforms them across all tasks while maintaining the same computational budget. This makes MTU an ideal candidate while designing multi-task models for resource constraint settings. Based on our findings, we propose an improved recipe for enabling multi-task capabilities in pre-trained models. 

\begin{itemize}
    \item Start by adding task-specific input convolution layers to process additional image conditioning. Introduce task-specific layer norms before FFNs, and fine-tune the model without splitting any FFN into smaller experts (i.e., keeping a single expert). As shown in Table \ref{tab:ablation-results}, this simple approach performs well for SDXL but is less effective for SDv1.5.
    \item If a single FFN does not yield satisfactory performance, conduct an ablation study over the number of experts to determine the optimal number of experts for improved multi-task performance.
\end{itemize}



\section{Conclusions}

In this work we introduced Multi-task Upcycling, a simple yet effective approach to enhance pre-trained text-to-image models, such as SDv1.5 and SDXL, to support multiple image editing tasks. Unlike previous approaches, MTU is the first multi-task diffusion modeling framework that seamlessly integrates multi-task learning with on-device compatibility, ensuring efficiency without compromising performance. Our idea is based on an empirical observation that parameters in FFN layers in diffusion models deviate the most during task-specific fine-tuning. We use this observation to propose splitting of existing FFN layer into smaller FFN experts, which are then combined with a router network. We conduct an extensive evaluation across Text-to-Image, Image Editing, Super-Resolution, and Inpainting tasks, demonstrating superior performance compared to both single-task and multi-task baselines. We believe that our approach will open up new avenues of research in the rapidly evolving area of image synthesis and will continue helping the efforts in making multi-task vision models efficient for on-device deployment.

\bibliography{main}

\begin{thebibliography}{60}
\providecommand{\natexlab}[1]{#1}
\providecommand{\url}[1]{\texttt{#1}}
\expandafter\ifx\csname urlstyle\endcsname\relax
  \providecommand{\doi}[1]{doi: #1}\else
  \providecommand{\doi}{doi: \begingroup \urlstyle{rm}\Url}\fi

\bibitem[Bao et~al.(2023)Bao, Nie, Xue, Li, Pu, Wang, Yue, Cao, Su, and Zhu]{bao2023one}
Bao, F., Nie, S., Xue, K., Li, C., Pu, S., Wang, Y., Yue, G., Cao, Y., Su, H., and Zhu, J.
\newblock One transformer fits all distributions in multi-modal diffusion at scale.
\newblock In \emph{International Conference on Machine Learning}, pp.\  1692--1717. PMLR, 2023.

\bibitem[{Brooks} et~al.(2022){Brooks}, {Holynski}, and {Efros}]{brooks2023instructpix2pix}
{Brooks}, T., {Holynski}, A., and {Efros}, A.~A.
\newblock {InstructPix2Pix}: Learning to follow image editing instructions.
\newblock In \emph{\href{https://arxiv.org/abs/2211.09800}{arXiv:2211.09800}}, 2022.

\bibitem[Brooks et~al.(2023)Brooks, Holynski, and Efros]{brooks2022instructpix2pix}
Brooks, T., Holynski, A., and Efros, A.~A.
\newblock Instructpix2pix: Learning to follow image editing instructions.
\newblock In \emph{CVPR}, 2023.

\bibitem[Castells et~al.(2024)Castells, Song, Piao, Choi, Kim, Yim, Lee, Kim, and Kim]{castells2024edgefusion}
Castells, T., Song, H.-K., Piao, T., Choi, S., Kim, B.-K., Yim, H., Lee, C., Kim, J.~G., and Kim, T.-H.
\newblock Edgefusion: On-device text-to-image generation.
\newblock \emph{CVPR24 First Workshop on Efficient and On-Device Generation (EDGE)}, 2024.

\bibitem[Chen et~al.(2024)Chen, Ding, Sisman, Xu, Xie, Yao, Tran, and Zeng]{chen2024diffusion}
Chen, C., Ding, H., Sisman, B., Xu, Y., Xie, O., Yao, B.~Z., Tran, S.~D., and Zeng, B.
\newblock Diffusion models for multi-task generative modeling.
\newblock In \emph{The Twelfth International Conference on Learning Representations}, 2024.
\newblock URL \url{https://openreview.net/forum?id=cbv0sBIZh9}.

\bibitem[Clark et~al.(2022)Clark, Casas, Guy, Mensch, Paganini, Hoffmann, Damoc, Hechtman, Cai, Borgeaud, Driessche, Rutherford, Hennigan, Johnson, Millican, Cassirer, Jones, Buchatskaya, Budden, Sifre, Osindero, Vinyals, Rae, Elsen, Kavukcuoglu, and Simonyan]{clark2022unified}
Clark, A., Casas, D. d.~l., Guy, A., Mensch, A., Paganini, M., Hoffmann, J., Damoc, B., Hechtman, B., Cai, T., Borgeaud, S., Driessche, G. v.~d., Rutherford, E., Hennigan, T., Johnson, M., Millican, K., Cassirer, A., Jones, C., Buchatskaya, E., Budden, D., Sifre, L., Osindero, S., Vinyals, O., Rae, J., Elsen, E., Kavukcuoglu, K., and Simonyan, K.
\newblock Unified scaling laws for routed language models.
\newblock In \emph{Proceedings of the 39th International Conference on Machine Learning}. PMLR, 2022.

\bibitem[Corneanu et~al.(2024)Corneanu, Gadde, and Martinez]{Corneanu2024Inpaint}
Corneanu, C., Gadde, R., and Martinez, A.~M.
\newblock Latentpaint: Image inpainting in latent space with diffusion models.
\newblock In \emph{2024 IEEE/CVF Winter Conference on Applications of Computer Vision (WACV)}, pp.\  4322--4331, 2024.
\newblock \doi{10.1109/WACV57701.2024.00428}.

\bibitem[Esser et~al.(2024)Esser, Kulal, Blattmann, Entezari, Muller, Saini, Levi, Lorenz, Sauer, Boesel, Podell, Dockhorn, English, Lacey, Goodwin, Marek, and Rombach]{Esser2024ScalingRF}
Esser, P., Kulal, S., Blattmann, A., Entezari, R., Muller, J., Saini, H., Levi, Y., Lorenz, D., Sauer, A., Boesel, F., Podell, D., Dockhorn, T., English, Z., Lacey, K., Goodwin, A., Marek, Y., and Rombach, R.
\newblock Scaling rectified flow transformers for high-resolution image synthesis.
\newblock \emph{ArXiv}, abs/2403.03206, 2024.
\newblock URL \url{https://api.semanticscholar.org/CorpusID:268247980}.

\bibitem[Fang et~al.(2023)Fang, Ma, and Wang]{fang2023structural}
Fang, G., Ma, X., and Wang, X.
\newblock Structural pruning for diffusion models.
\newblock In \emph{Advances in Neural Information Processing Systems}, 2023.

\bibitem[Gao et~al.(2024)Gao, Zhuo, Liu, , Du, Luo, Qiu, Zhang, et~al.]{gao2024lumin-t2x}
Gao, P., Zhuo, L., Liu, C., , Du, R., Luo, X., Qiu, L., Zhang, Y., et~al.
\newblock Lumina-t2x: Transforming text into any modality, resolution, and duration via flow-based large diffusion transformers.
\newblock \emph{arXiv preprint arXiv:2405.05945}, 2024.

\bibitem[He et~al.(2024)He, Khattar, Prenger, Korthikanti, Yan, Liu, Fan, Aithal, Shoeybi, and Catanzaro]{he2024upcyclinglargelanguagemodels}
He, E., Khattar, A., Prenger, R., Korthikanti, V., Yan, Z., Liu, T., Fan, S., Aithal, A., Shoeybi, M., and Catanzaro, B.
\newblock Upcycling large language models into mixture of experts, 2024.

\bibitem[He et~al.(2023)He, Liu, Liu, Wu, Zhou, and Zhuang]{he2023ptqd}
He, Y., Liu, L., Liu, J., Wu, W., Zhou, H., and Zhuang, B.
\newblock {PTQD}: Accurate post-training quantization for diffusion models.
\newblock In \emph{Thirty-seventh Conference on Neural Information Processing Systems}, 2023.
\newblock URL \url{https://openreview.net/forum?id=Y3g1PV5R9l}.

\bibitem[Ho \& Salimans(2022)Ho and Salimans]{ho2022classifierfreediffusionguidance}
Ho, J. and Salimans, T.
\newblock Classifier-free diffusion guidance.
\newblock \emph{NeurIPS 2021 Workshop on Deep Generative Models and Downstream Applications}, 2022.

\bibitem[Ho et~al.(2020)Ho, Jain, and Abbeel]{ho2020denoising}
Ho, J., Jain, A., and Abbeel, P.
\newblock Denoising diffusion probabilistic models.
\newblock \emph{arXiv preprint arxiv:2006.11239}, 2020.

\bibitem[Hu et~al.(2022)Hu, Shen, Wallis, Allen-Zhu, Li, Wang, Wang, and Chen]{hu2022lora}
Hu, E.~J., Shen, Y., Wallis, P., Allen-Zhu, Z., Li, Y., Wang, S., Wang, L., and Chen, W.
\newblock Lo{RA}: Low-rank adaptation of large language models.
\newblock In \emph{International Conference on Learning Representations}, 2022.
\newblock URL \url{https://openreview.net/forum?id=nZeVKeeFYf9}.

\bibitem[Hudson \& Manning(2019)Hudson and Manning]{8953451}
Hudson, D.~A. and Manning, C.~D.
\newblock Gqa: A new dataset for real-world visual reasoning and compositional question answering.
\newblock In \emph{2019 IEEE/CVF Conference on Computer Vision and Pattern Recognition (CVPR)}, 2019.

\bibitem[Jiang et~al.(2025)Jiang, Lu, Lin, Han, and Sun]{jiang-etal-2025-improved}
Jiang, W., Lu, Y., Lin, H., Han, X., and Sun, L.
\newblock Improved sparse upcycling for instruction tuning.
\newblock In \emph{Proceedings of the 31st International Conference on Computational Linguistics}, 2025.

\bibitem[Kang et~al.(2025)Kang, Zhang, Barnes, Paris, Kwak, Park, Shechtman, Zhu, and Park]{kang2025distilling}
Kang, M., Zhang, R., Barnes, C., Paris, S., Kwak, S., Park, J., Shechtman, E., Zhu, J.-Y., and Park, T.
\newblock Distilling diffusion models into conditional gans.
\newblock In \emph{European Conference on Computer Vision}, pp.\  428--447. Springer, 2025.

\bibitem[Kirkpatrick et~al.(2017)Kirkpatrick, Pascanu, Rabinowitz, Veness, Desjardins, Rusu, Milan, Quan, Ramalho, Grabska-Barwinska, Hassabis, Clopath, Kumaran, and Hadsell]{overcoming2017james}
Kirkpatrick, J., Pascanu, R., Rabinowitz, N., Veness, J., Desjardins, G., Rusu, A.~A., Milan, K., Quan, J., Ramalho, T., Grabska-Barwinska, A., Hassabis, D., Clopath, C., Kumaran, D., and Hadsell, R.
\newblock Overcoming catastrophic forgetting in neural networks.
\newblock \emph{Proceedings of the National Academy of Sciences}, 2017.

\bibitem[Komatsuzaki et~al.(2023)Komatsuzaki, Puigcerver, Lee-Thorp, Ruiz, Mustafa, Ainslie, Tay, Dehghani, and Houlsby]{komatsuzaki2023sparseupcyclingtrainingmixtureofexperts}
Komatsuzaki, A., Puigcerver, J., Lee-Thorp, J., Ruiz, C.~R., Mustafa, B., Ainslie, J., Tay, Y., Dehghani, M., and Houlsby, N.
\newblock Sparse upcycling: Training mixture-of-experts from dense checkpoints, 2023.

\bibitem[Krajewski et~al.(2024)Krajewski, Ludziejewski, Adamczewski, Pi'oro, Krutul, Antoniak, Ciebiera, Kr'ol, Odrzyg'o'zd'z, Sankowski, Cygan, and Jaszczur]{Krajewski2024ScalingLF}
Krajewski, J., Ludziejewski, J., Adamczewski, K., Pi'oro, M., Krutul, M., Antoniak, S., Ciebiera, K., Kr'ol, K., Odrzyg'o'zd'z, T., Sankowski, P., Cygan, M., and Jaszczur, S.
\newblock Scaling laws for fine-grained mixture of experts.
\newblock \emph{ArXiv}, abs/2402.07871, 2024.
\newblock URL \url{https://api.semanticscholar.org/CorpusID:267626982}.

\bibitem[Lewis et~al.(2021)Lewis, Bhosale, Dettmers, Goyal, and Zettlemoyer]{pmlr-v139-lewis21a}
Lewis, M., Bhosale, S., Dettmers, T., Goyal, N., and Zettlemoyer, L.
\newblock Base layers: Simplifying training of large, sparse models.
\newblock In \emph{Proceedings of the 38th International Conference on Machine Learning}. PMLR, 2021.

\bibitem[Li et~al.(2017)Li, Kadav, Durdanovic, Samet, and Graf]{li2017pruning}
Li, H., Kadav, A., Durdanovic, I., Samet, H., and Graf, H.~P.
\newblock Pruning filters for efficient convnets.
\newblock In \emph{International Conference on Learning Representations}, 2017.
\newblock URL \url{https://openreview.net/forum?id=rJqFGTslg}.

\bibitem[Li et~al.(2023)Li, Liu, Lian, Yang, Dong, Kang, Zhang, and Keutzer]{li2023qdiffusion}
Li, X., Liu, Y., Lian, L., Yang, H., Dong, Z., Kang, D., Zhang, S., and Keutzer, K.
\newblock Q-diffusion: Quantizing diffusion models.
\newblock In \emph{Proceedings of the IEEE/CVF International Conference on Computer Vision (ICCV)}, pp.\  17535--17545, October 2023.

\bibitem[{Lin} et~al.(2024){Lin}, {Wang}, and {Yang}]{2024arXiv240213929L_sdxl-lightning}
{Lin}, S., {Wang}, A., and {Yang}, X.
\newblock {SDXL-Lightning}: Progressive adversarial diffusion distillation.
\newblock In \emph{\href{https://arxiv.org/abs/2402.13929}{arXiv:2402.13929}}, 2024.

\bibitem[Lin et~al.(2015)Lin, Maire, Belongie, Bourdev, Girshick, Hays, Perona, Ramanan, Zitnick, and Dollár]{lin2015microsoftcococommonobjects}
Lin, T.-Y., Maire, M., Belongie, S., Bourdev, L., Girshick, R., Hays, J., Perona, P., Ramanan, D., Zitnick, C.~L., and Dollár, P.
\newblock Microsoft coco: Common objects in context.
\newblock \emph{arXiv preprint arXiv:2405.19237}, 2015.

\bibitem[Liu et~al.(2022)Liu, Tam, Mohammed, Mohta, Huang, Bansal, and Raffel]{liu2022fewshot}
Liu, H., Tam, D., Mohammed, M., Mohta, J., Huang, T., Bansal, M., and Raffel, C.
\newblock Few-shot parameter-efficient fine-tuning is better and cheaper than in-context learning.
\newblock In Oh, A.~H., Agarwal, A., Belgrave, D., and Cho, K. (eds.), \emph{Advances in Neural Information Processing Systems}, 2022.
\newblock URL \url{https://openreview.net/forum?id=rBCvMG-JsPd}.

\bibitem[Loshchilov \& Hutter(2017)Loshchilov and Hutter]{loshchilov2017sgdr}
Loshchilov, I. and Hutter, F.
\newblock {SGDR}: Stochastic gradient descent with warm restarts.
\newblock In \emph{International Conference on Learning Representations}, 2017.
\newblock URL \url{https://openreview.net/forum?id=Skq89Scxx}.

\bibitem[Meng et~al.(2023)Meng, Rombach, Gao, Kingma, Ermon, Ho, and Salimans]{meng2023distillation}
Meng, C., Rombach, R., Gao, R., Kingma, D., Ermon, S., Ho, J., and Salimans, T.
\newblock On distillation of guided diffusion models.
\newblock In \emph{Proceedings of the IEEE/CVF Conference on Computer Vision and Pattern Recognition}, pp.\  14297--14306, 2023.

\bibitem[Moser et~al.(2024)Moser, Shanbhag, Raue, Frolov, Palacio, and Dengel]{Moser_2024}
Moser, B.~B., Shanbhag, A.~S., Raue, F., Frolov, S., Palacio, S., and Dengel, A.
\newblock Diffusion models, image super-resolution, and everything: A survey.
\newblock \emph{IEEE Transactions on Neural Networks and Learning Systems}, pp.\  1–21, 2024.
\newblock ISSN 2162-2388.
\newblock \doi{10.1109/tnnls.2024.3476671}.
\newblock URL \url{http://dx.doi.org/10.1109/TNNLS.2024.3476671}.

\bibitem[Noroozi et~al.(2024)Noroozi, Hadji, Martinez, Bulat, and Tzimiropoulos]{noroozi2024needstepfastsuperresolution}
Noroozi, M., Hadji, I., Martinez, B., Bulat, A., and Tzimiropoulos, G.
\newblock You only need one step: Fast super-resolution with stable diffusion via scale distillation, 2024.
\newblock URL \url{https://arxiv.org/abs/2401.17258}.

\bibitem[{Peebles} \& {Xie}(2022){Peebles} and {Xie}]{2022arXiv221209748P_dit}
{Peebles}, W. and {Xie}, S.
\newblock Scalable diffusion models with transformers.
\newblock In \emph{\href{https://arxiv.org/abs/2212.09748}{arXiv:2212.09748}}, 2022.

\bibitem[{Podell} et~al.(2023){Podell}, {English}, {Lacey}, {Blattmann}, {Dockhorn}, {M{\"u}ller}, {Penna}, and {Rombach}]{2023arXiv230701952P_sdxl}
{Podell}, D., {English}, Z., {Lacey}, K., {Blattmann}, A., {Dockhorn}, T., {M{\"u}ller}, J., {Penna}, J., and {Rombach}, R.
\newblock {SDXL}: Improving latent diffusion models for high-resolution image synthesis.
\newblock In \emph{\href{https://arxiv.org/abs/2307.01952}{arXiv:2307.01952}}, 2023.

\bibitem[Radford et~al.(2021)Radford, Kim, Hallacy, Ramesh, Goh, Agarwal, Sastry, Askell, Mishkin, Clark, Krueger, and Sutskever]{radford2021learningtransferablevisualmodels}
Radford, A., Kim, J.~W., Hallacy, C., Ramesh, A., Goh, G., Agarwal, S., Sastry, G., Askell, A., Mishkin, P., Clark, J., Krueger, G., and Sutskever, I.
\newblock Learning transferable visual models from natural language supervision.
\newblock \emph{arXiv}, 2021.

\bibitem[Ramesh et~al.(2022)Ramesh, Prafulla, Alex, Casey, and Mark]{ramesh2022}
Ramesh, A., Prafulla, D., Alex, N., Casey, C., and Mark, C.
\newblock Hierarchical text-conditional image generation with clip latents.
\newblock 2022.

\bibitem[Rombach et~al.(2022)Rombach, Blattmann, Lorenz, Esser, and Ommer]{2022CVPRLDM}
Rombach, R., Blattmann, A., Lorenz, D., Esser, P., and Ommer, B.
\newblock {High-resolution image synthesis with latent diffusion models}.
\newblock In \emph{\href{https://openaccess.thecvf.com/content/CVPR2022/html/Rombach_High-Resolution_Image_Synthesis_With_Latent_Diffusion_Models_CVPR_2022_paper.html}{CVPR}}, 2022.

\bibitem[Salimans \& Ho(2022)Salimans and Ho]{salimans2022progressive}
Salimans, T. and Ho, J.
\newblock Progressive distillation for fast sampling of diffusion models.
\newblock In \emph{International Conference on Learning Representations}, 2022.
\newblock URL \url{https://openreview.net/forum?id=TIdIXIpzhoI}.

\bibitem[Shazeer et~al.(2017)Shazeer, Mirhoseini, Maziarz, Davis, Le, Hinton, and Dean]{Shazeer2017OutrageouslyLN}
Shazeer, N.~M., Mirhoseini, A., Maziarz, K., Davis, A., Le, Q.~V., Hinton, G.~E., and Dean, J.
\newblock Outrageously large neural networks: The sparsely-gated mixture-of-experts layer.
\newblock \emph{ArXiv}, abs/1701.06538, 2017.
\newblock URL \url{https://api.semanticscholar.org/CorpusID:12462234}.

\bibitem[Song et~al.(2021)Song, Meng, and Ermon]{song2021denoising}
Song, J., Meng, C., and Ermon, S.
\newblock Denoising diffusion implicit models.
\newblock In \emph{International Conference on Learning Representations}, 2021.
\newblock URL \url{https://openreview.net/forum?id=St1giarCHLP}.

\bibitem[{Stability AI}(2023)]{stability_ai_sdxl_turbo}
{Stability AI}.
\newblock {SDXL-Turbo}.
\newblock \url{https://huggingface.co/stabilityai/sdxl-turbo}, 2023.

\bibitem[Tang et~al.(2023)Tang, Wang, Chen, Guan, Tang, and zhu]{tang2023lightweightdiffusionmodelsdistillationbased}
Tang, S., Wang, X., Chen, H., Guan, C., Tang, Y., and zhu, W.
\newblock Lightweight diffusion models with distillation-based block neural architecture search, 2023.
\newblock URL \url{https://arxiv.org/abs/2311.04950}.

\bibitem[Tang et~al.(2024)Tang, Yang, Zhu, Zeng, and Bansal]{tang2024any}
Tang, Z., Yang, Z., Zhu, C., Zeng, M., and Bansal, M.
\newblock Any-to-any generation via composable diffusion.
\newblock \emph{Advances in Neural Information Processing Systems}, 36, 2024.

\bibitem[Wang et~al.(2024)Wang, Shang, Yuan, Wu, and Yan]{wang2024quest}
Wang, H., Shang, Y., Yuan, Z., Wu, J., and Yan, Y.
\newblock Quest: Low-bit diffusion model quantization via efficient selective finetuning, 2024.

\bibitem[Wang et~al.(2021)Wang, Xie, Dong, and Shan]{wang2021real}
Wang, X., Xie, L., Dong, C., and Shan, Y.
\newblock Real-esrgan: Training real-world blind super-resolution with pure synthetic data.
\newblock In \emph{Proceedings of the IEEE/CVF international conference on computer vision}, pp.\  1905--1914, 2021.

\bibitem[Wasserman et~al.(2024)Wasserman, Rotstein, Ganz, and Kimmel]{wasserman2024paint}
Wasserman, N., Rotstein, N., Ganz, R., and Kimmel, R.
\newblock Paint by inpaint: Learning to add image objects by removing them first, 2024.

\bibitem[Xiang et~al.(2024)Xiang, Zhang, Shang, Wu, Yan, and Nie]{xiang2024dkdmdatafreeknowledgedistillation}
Xiang, Q., Zhang, M., Shang, Y., Wu, J., Yan, Y., and Nie, L.
\newblock Dkdm: Data-free knowledge distillation for diffusion models with any architecture, 2024.
\newblock URL \url{https://arxiv.org/abs/2409.03550}.

\bibitem[Xie et~al.(2024)Xie, Chen, Chen, Cai, Tang, Lin, Zhang, Li, Zhu, Lu, and Han]{xie2024sana}
Xie, E., Chen, J., Chen, J., Cai, H., Tang, H., Lin, Y., Zhang, Z., Li, M., Zhu, L., Lu, Y., and Han, S.
\newblock Sana: Efficient high-resolution image synthesis with linear diffusion transformer, 2024.
\newblock URL \url{https://arxiv.org/abs/2410.10629}.

\bibitem[Xu et~al.(2023)Xu, Wang, Zhang, Wang, and Shi]{xu2022versatile}
Xu, X., Wang, Z., Zhang, G., Wang, K., and Shi, H.
\newblock Versatile diffusion: Text, images and variations all in one diffusion model.
\newblock pp.\  7754--7765, 2023.

\bibitem[Ye \& Xu(2024)Ye and Xu]{diffusionmtl}
Ye, H. and Xu, D.
\newblock Diffusionmtl: Learning multi-task denoising diffusion model from partially annotated data.
\newblock In \emph{CVPR}, 2024.

\bibitem[Yildirim et~al.(2023)Yildirim, Baday, Erdem, Erdem, and Dundar]{yildirim2023instinpaint}
Yildirim, A.~B., Baday, V., Erdem, E., Erdem, A., and Dundar, A.
\newblock Inst-inpaint: Instructing to remove objects with diffusion models, 2023.

\bibitem[Yin et~al.(2024)Yin, Gharbi, Zhang, Shechtman, Durand, Freeman, and Park]{yin2024one}
Yin, T., Gharbi, M., Zhang, R., Shechtman, E., Durand, F., Freeman, W.~T., and Park, T.
\newblock One-step diffusion with distribution matching distillation.
\newblock In \emph{Proceedings of the IEEE/CVF Conference on Computer Vision and Pattern Recognition}, pp.\  6613--6623, 2024.

\bibitem[Zhang et~al.(2024{\natexlab{a}})Zhang, Ye, Li, Wang, and Xu]{zhang2024multi}
Zhang, J., Ye, H., Li, X., Wang, W., and Xu, D.
\newblock Multi-task label discovery via hierarchical task tokens for partially annotated dense predictions.
\newblock \emph{arXiv preprint arXiv:2411.18823}, 2024{\natexlab{a}}.

\bibitem[Zhang et~al.(2023)Zhang, Rao, and Agrawala]{Zhang2023ICCV}
Zhang, L., Rao, A., and Agrawala, M.
\newblock Adding conditional control to text-to-image diffusion models.
\newblock In \emph{Proceedings of the IEEE/CVF International Conference on Computer Vision (ICCV)}, pp.\  3836--3847, October 2023.

\bibitem[Zhang et~al.(2018)Zhang, Isola, Efros, Shechtman, and Wang]{zhang2018perceptual}
Zhang, R., Isola, P., Efros, A.~A., Shechtman, E., and Wang, O.
\newblock The unreasonable effectiveness of deep features as a perceptual metric.
\newblock In \emph{CVPR}, 2018.

\bibitem[Zhang et~al.(2024{\natexlab{b}})Zhang, Zhang, Yue, Lu, Ren, and Shen]{zhang2024mobile}
Zhang, Y., Zhang, J., Yue, S., Lu, W., Ren, J., and Shen, X.
\newblock Mobile generative ai: Opportunities and challenges.
\newblock \emph{IEEE Wireless Communications}, 31\penalty0 (4):\penalty0 58--64, 2024{\natexlab{b}}.

\bibitem[{Zhao} et~al.(2023){Zhao}, {Xu}, {Xiao}, and {Hou}]{2023arXiv231116567Z_mobilediffusion}
{Zhao}, Y., {Xu}, Y., {Xiao}, Z., and {Hou}, T.
\newblock {MobileDiffusion}: Subsecond text-to-image generation on mobile devices.
\newblock In \emph{\href{https://arxiv.org/abs/2311.16567}{arXiv:2311.16567}}, 2023.

\bibitem[Zhao et~al.(2025)Zhao, Xu, Xiao, Jia, and Hou]{ZhaoXXJH24}
Zhao, Y., Xu, Y., Xiao, Z., Jia, H., and Hou, T.
\newblock Mobilediffusion: Instant text-to-image generation on mobile devices.
\newblock In \emph{European Conference on Computer Vision (ECCV)}, pp.\  225--242. Springer, 2025.

\bibitem[Zhou et~al.(2022)Zhou, Lei, Liu, Du, Huang, Zhao, Dai, Chen, Le, and Laudon]{Zhou2022MixtureofExpertsWE}
Zhou, Y.-Q., Lei, T., Liu, H.-C., Du, N., Huang, Y., Zhao, V., Dai, A.~M., Chen, Z., Le, Q.~V., and Laudon, J.
\newblock Mixture-of-experts with expert choice routing.
\newblock \emph{ArXiv}, abs/2202.09368, 2022.
\newblock URL \url{https://api.semanticscholar.org/CorpusID:247011948}.

\bibitem[Zhu et~al.(2025)Zhu, Liu, and Liu]{zhu2025slimflow}
Zhu, Y., Liu, X., and Liu, Q.
\newblock Slimflow: Training smaller one-step diffusion models with rectified flow.
\newblock In \emph{European Conference on Computer Vision}, pp.\  342--359. Springer, 2025.

\bibitem[Zhuo et~al.(2024)Zhuo, Du, Han, Li, Liu, Huang, Liu, et~al.]{gao2024lumina-next}
Zhuo, L., Du, R., Han, X., Li, Y., Liu, D., Huang, R., Liu, W., et~al.
\newblock Lumina-next: Making lumina-t2x stronger and faster with next-dit.
\newblock \emph{arXiv preprint arXiv:2406.18583}, 2024.

\end{thebibliography}
\bibliographystyle{icml2025}

\newpage
\appendix
\onecolumn

\section{Appendix}
\subsection{Dataset details}
\label{sec:app-datasets}
In this section, we provide details on the tasks and datasets used for training MTU models. Our study incorporates the following datasets, with the exact data splits outlined in Table \ref{tab:dataset-details}.

\begin{itemize}
\setlength{\itemsep}{1pt}
\setlength{\parskip}{0pt}
    \item Text-to-Image Generation (T2I): We utilize the COCO Captions dataset~\citep{lin2015microsoftcococommonobjects}, a large-scale collection of image-text pairs. Each image in this dataset is accompanied by five captions, with one randomly selected during training.
    \item Image Editing: We use the dataset introduced in \citep{brooks2022instructpix2pix}, which includes input-target image pairs along with corresponding editing instructions. Each input image has 4-5 target variations for a given edit instruction, with one randomly selected during training.
    \item Super Resolution: We use the Real-ESRGAN dataset \citep{wang2021real}, which consists of high-resolution images. We generate corresponding low-resolution images by applying degradations like Poisson and Gaussian blur and downscaling them by half.  For SR, we input an empty string to the model. 
    \item Image Inpainting: We utilize the dataset from \citep{yildirim2023instinpaint}, a multi-modal inpainting dataset designed for object removal based on text prompts. Built on the GQA dataset \citep{8953451}, it leverages scene graphs to generate paired training data using state-of-the-art instance segmentation and inpainting techniques.
\end{itemize}

\subsection{Training and Inference Details}
\label{sec:app-training}
In this section, we present training and inference details for both single-task and MTU models. 

\noindent
\textbf{Training details:} Both single-task and MTU models trained on 8× A100 GPUs for 100 epochs, with a batch size of 16 per GPU and image resolution of 512 $\times$ 512. SDXL is optimized using AdamW with a learning rate of 5e-5, while SDv1.5 is trained using Adam with a learning rate of 1e-4. For SDXL, we find that using a weight decay of 0.01 helps stabilize training. 

\noindent
\textbf{Inference details:} During sampling, we perform denoising for 20 iterations in multi-task SDv1.5 and 50 iterations in SDXL. For Text-to-Image (T2I) generation, Image Editing, and Inpainting, we apply Classifier-Free Guidance (CFG) \citep{ho2022classifierfreediffusionguidance}. However, for Super-Resolution (SR), no CFG is used, as it only processes an empty string as input. For T2I generation, we use a guidance scale of 7.5 for SDv1.5 and 5.0 for SDXL. For Image Editing and Inpainting, we follow the CFG strategy from \citep{brooks2022instructpix2pix}, which employs dual guidance scales—one for image and another for text. For SDv1.5, we use an image guidance scale of 1.6 and a text guidance scale of 7.5 for Image Editing, and 1.5 and 4.0 for Inpainting, respectively. For SDXL, we set the image guidance scale to 1.5 and the text guidance scale to 10.0 for Image Editing, while for Inpainting, we use 1.5 for image and 4.0 for text. For more details on the formulation of CFG, we direct the readers to \citep{brooks2022instructpix2pix}.

\begin{table}[h]
    \centering
    \begin{tabular}{|c|c|c|c|c|}
        \hline
         Task & Dataset & Train & Val & Test  \\
         \hline
         T2I & COCO Captions \citep{li2017pruning} & 118287 & 5000 & 5000 \\
         Image Editing & InstructPix2pix \citep{brooks2022instructpix2pix} & 281709 & 31301 & 2000 \\
         Super Resolution & Real- ESRGAN\citep{wang2021real} & 23744 & 100 & 100  \\
         Inpainting & GQA-Inpaint \citep{yildirim2023instinpaint} & 90089 & 10009 & 5553 \\
         \hline
    \end{tabular}
    \caption{Training, validation and test splits of the datasets used in training MTU}
    \label{tab:dataset-details}
\end{table}

\subsection{Qualitative Results}
\label{sec:app-pictures}
\begin{figure*}[] 
 \captionsetup[subfigure]{labelformat=empty}
    \centering
    \begin{subcaptionbox}{}{
\includegraphics[trim={{0cm} {3cm} {0cm} {0cm}}, clip,width=0.45\textwidth]{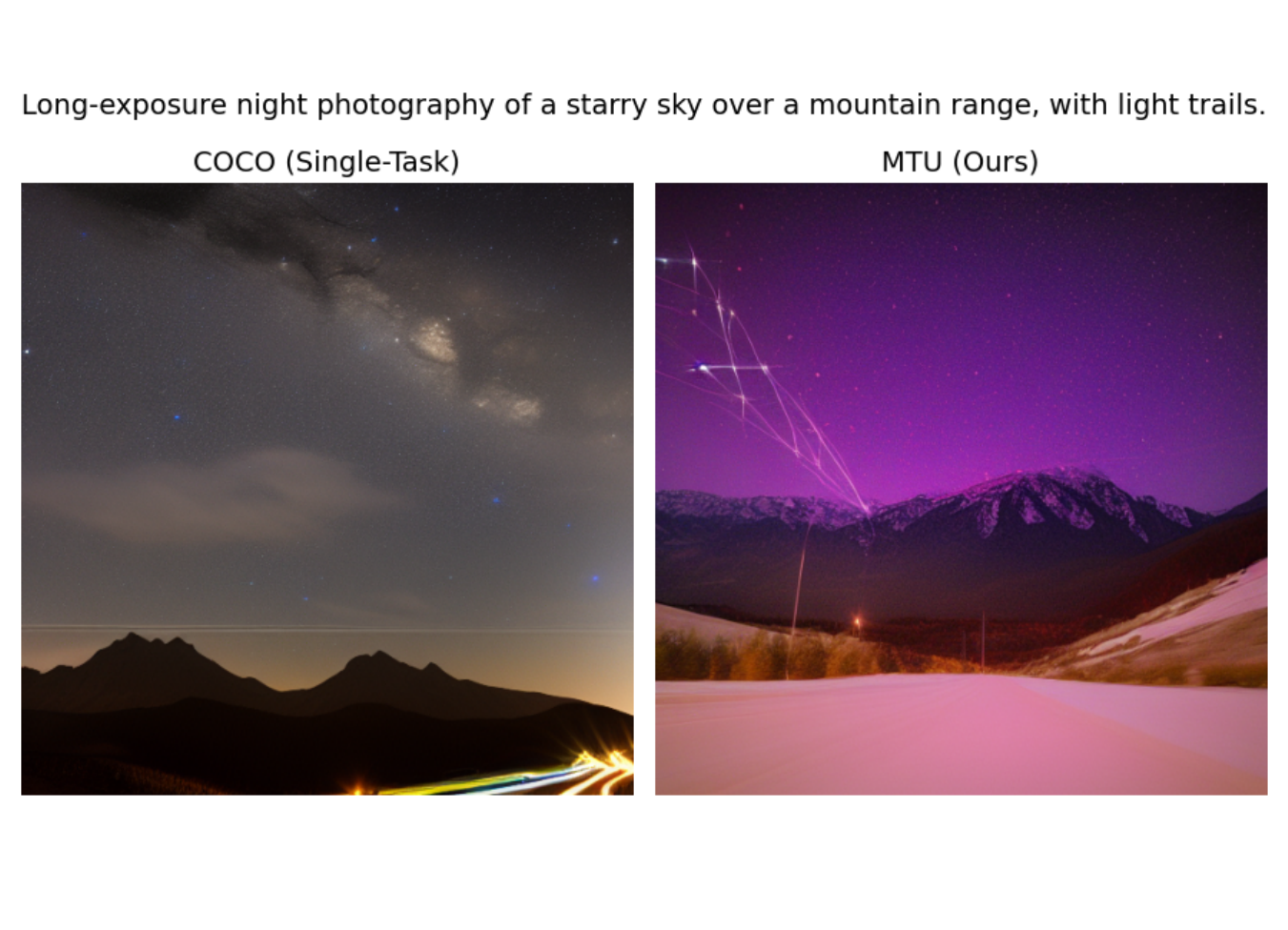}
    }\end{subcaptionbox}
    \vspace{-20pt}
    \begin{subcaptionbox}{}{
\includegraphics[trim={{0cm} {3cm} {0cm} {0cm}}, clip,width=0.45\textwidth]{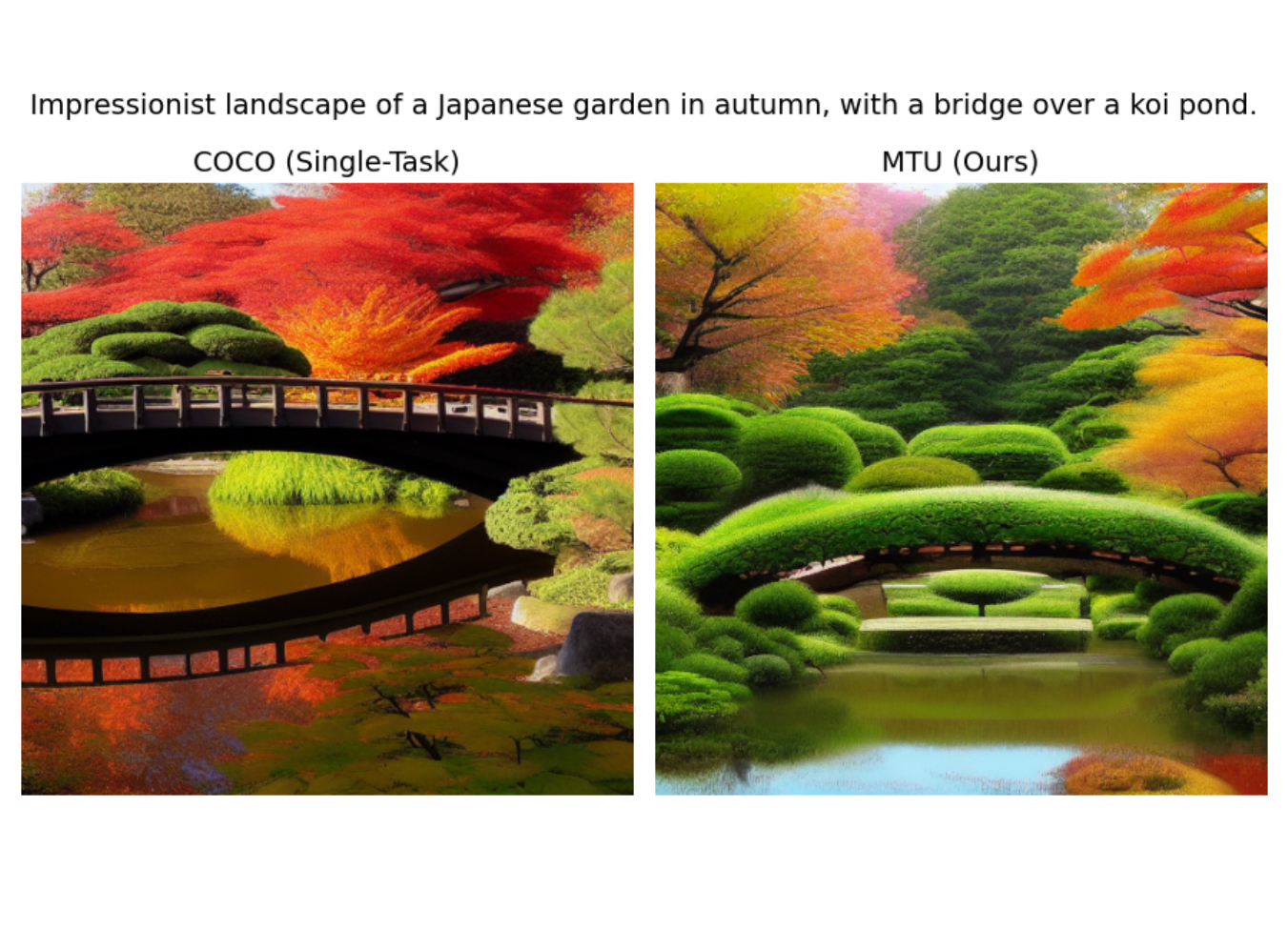}
    }\end{subcaptionbox}
        \begin{subcaptionbox}{}{
\includegraphics[trim={{0cm} {3cm} {0cm} {0cm}}, clip,width=0.45\textwidth]{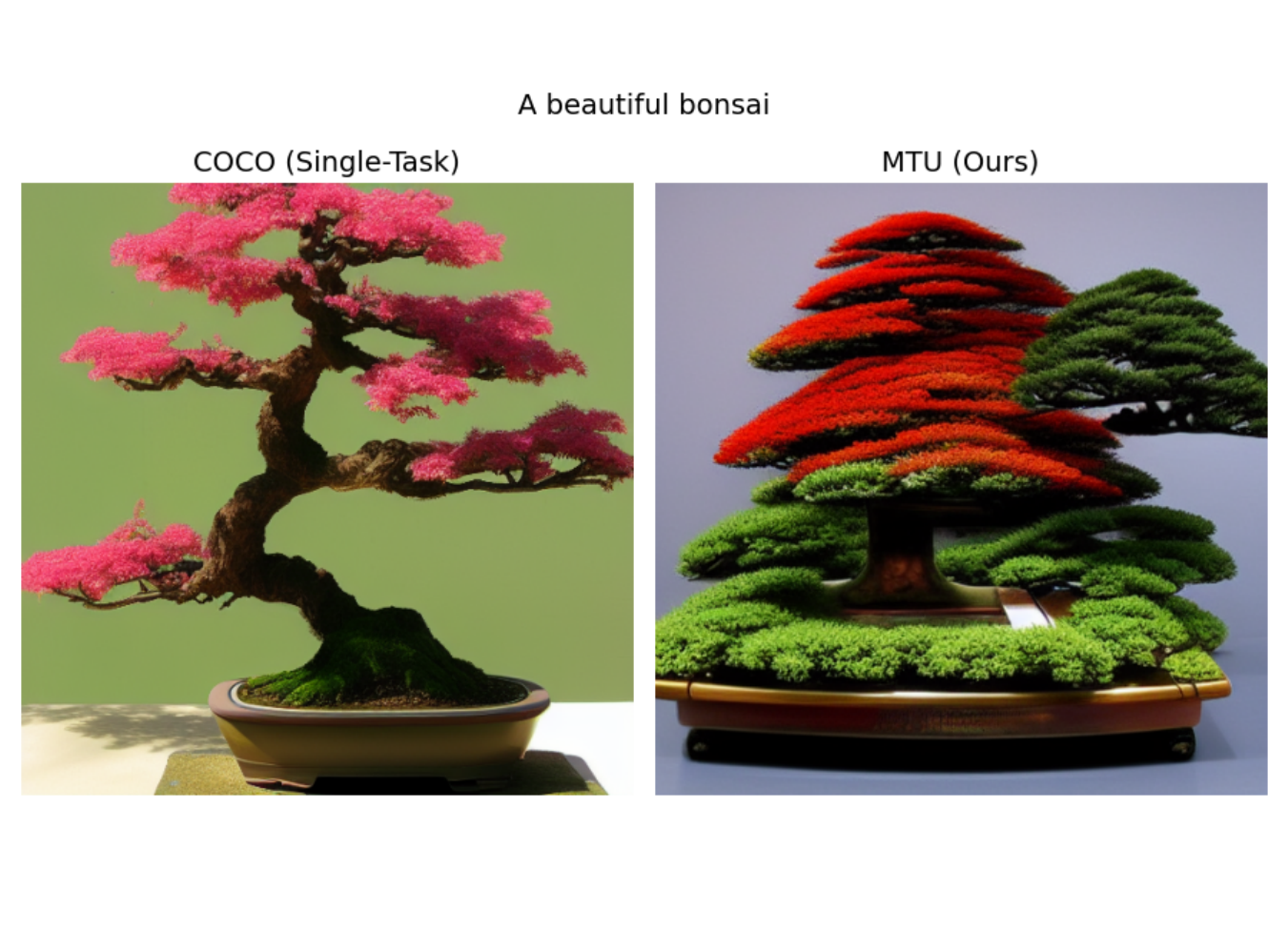}
    }\end{subcaptionbox}
    \vspace{-20pt}
    \begin{subcaptionbox}{}{
\includegraphics[trim={{0cm} {3cm} {0cm} {0cm}}, clip,width=0.45\textwidth]{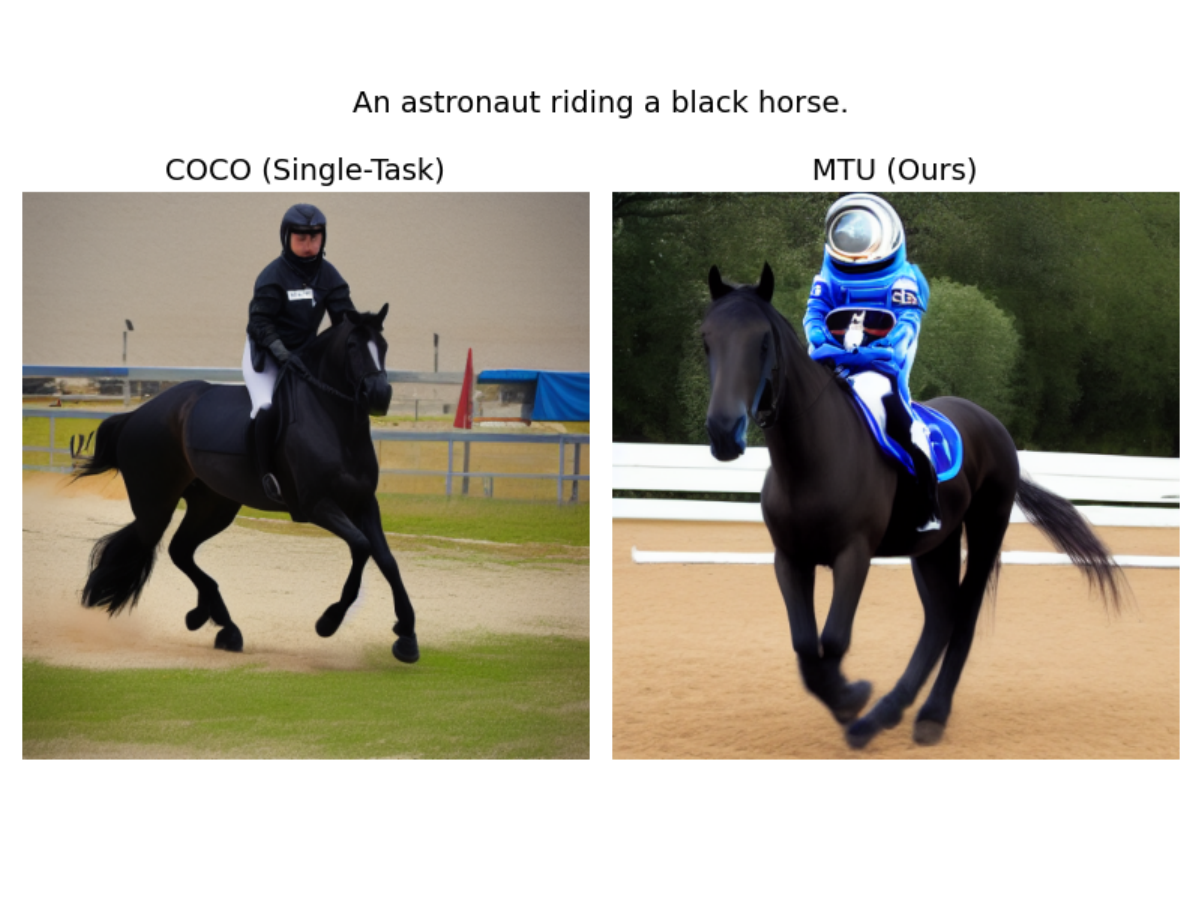}
    }\end{subcaptionbox}
       \begin{subcaptionbox}{}{
\includegraphics[trim={{0cm} {3cm} {0cm} {0cm}}, clip,width=0.45\textwidth]{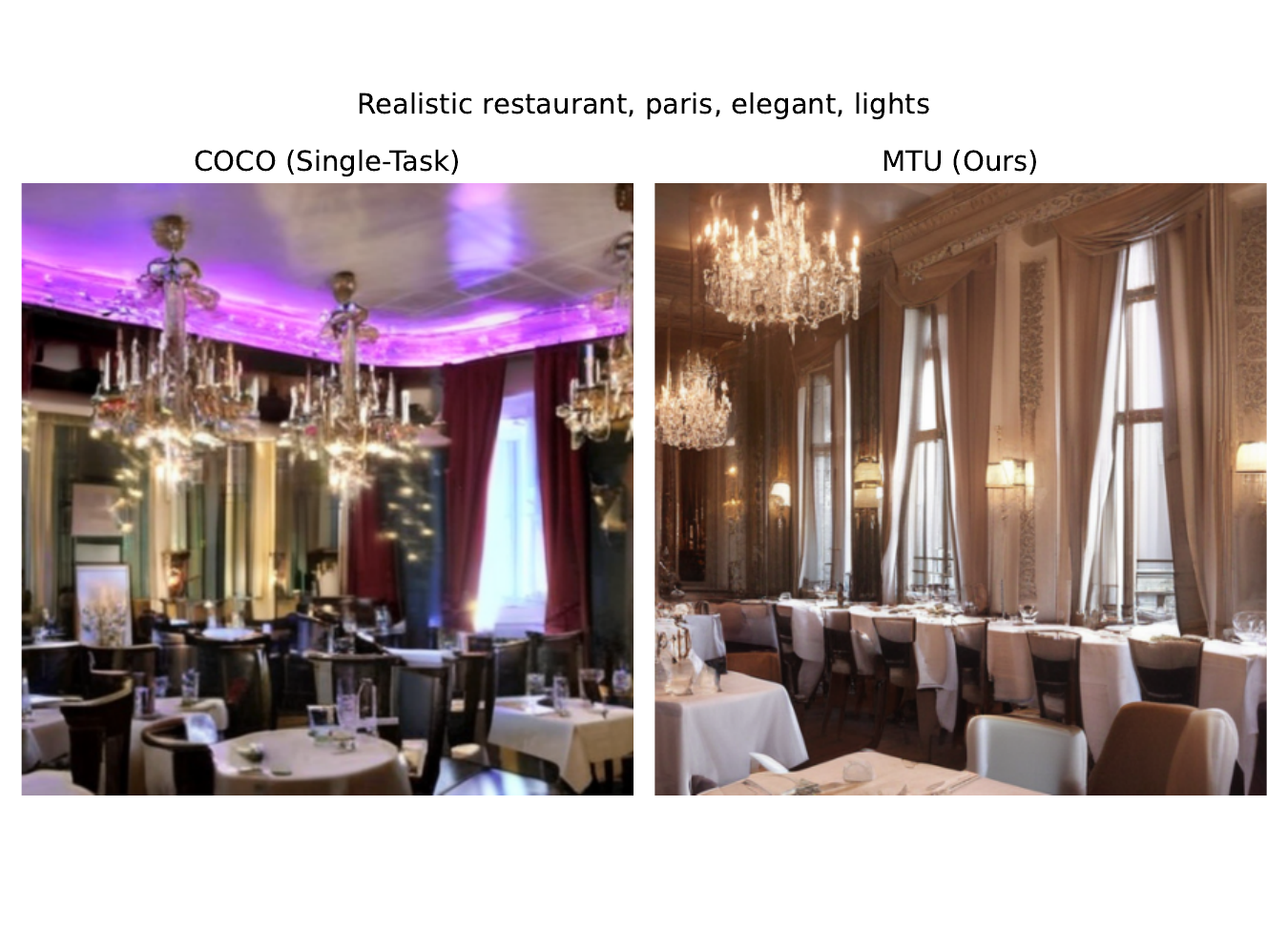}
    }\end{subcaptionbox}
    \vspace{-20pt}
    \begin{subcaptionbox}{}{
\includegraphics[trim={{0cm} {3cm} {0cm} {0cm}}, clip,width=0.45\textwidth]{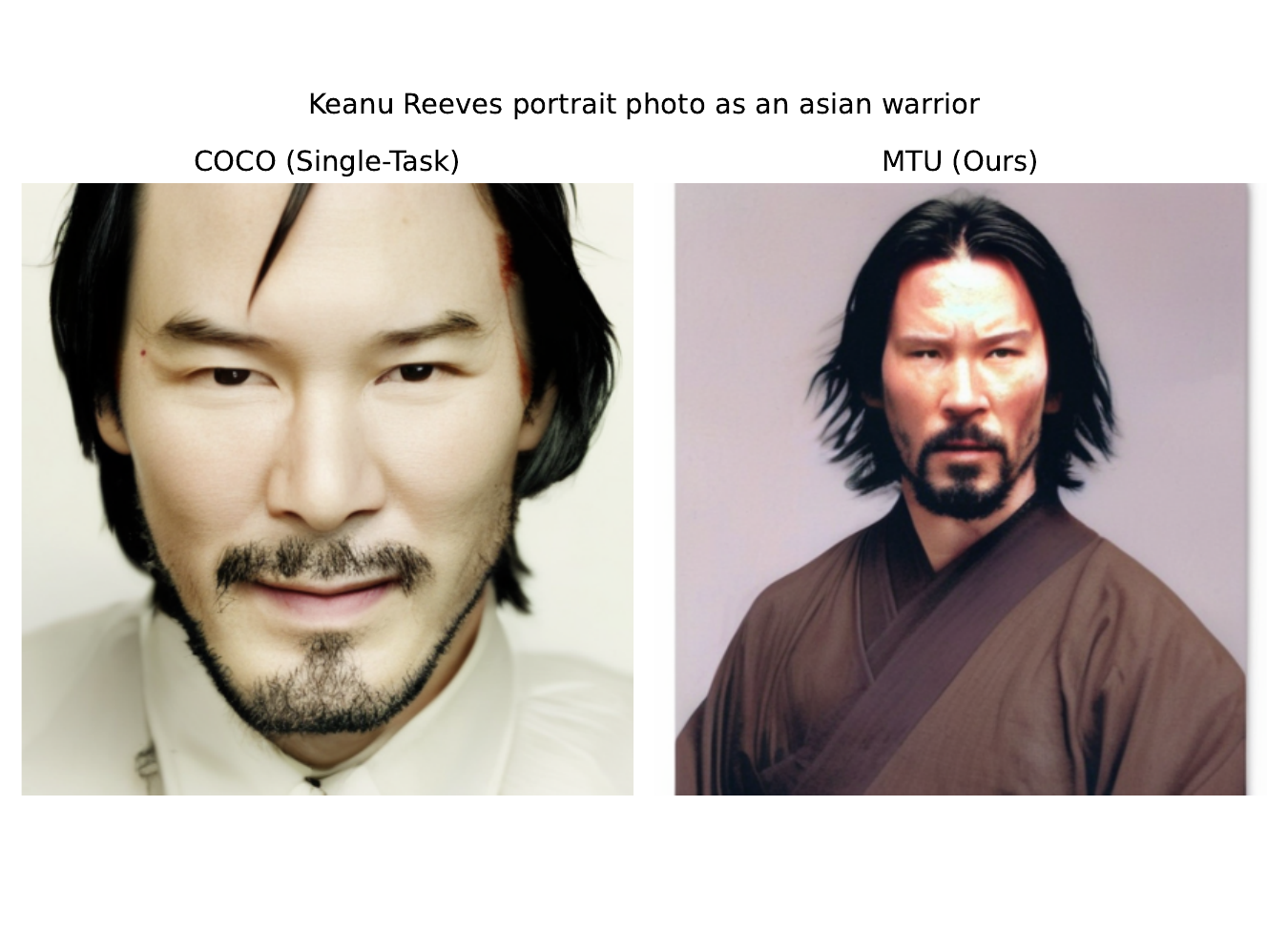}
    }\end{subcaptionbox}
    \caption{Qualitative results of MTU based on SDv1.5 (first two rows) and SDXL (bottom row) for Text-to-Image Generation.}
    \label{fig:pics-t2i}
\end{figure*}

\begin{figure*}[] 
 \captionsetup[subfigure]{labelformat=empty}
    \centering
    \begin{subcaptionbox}{}{
\begin{overpic}[trim={{0cm} {1cm} {0cm} {0cm}}, clip,width=0.45\textwidth]{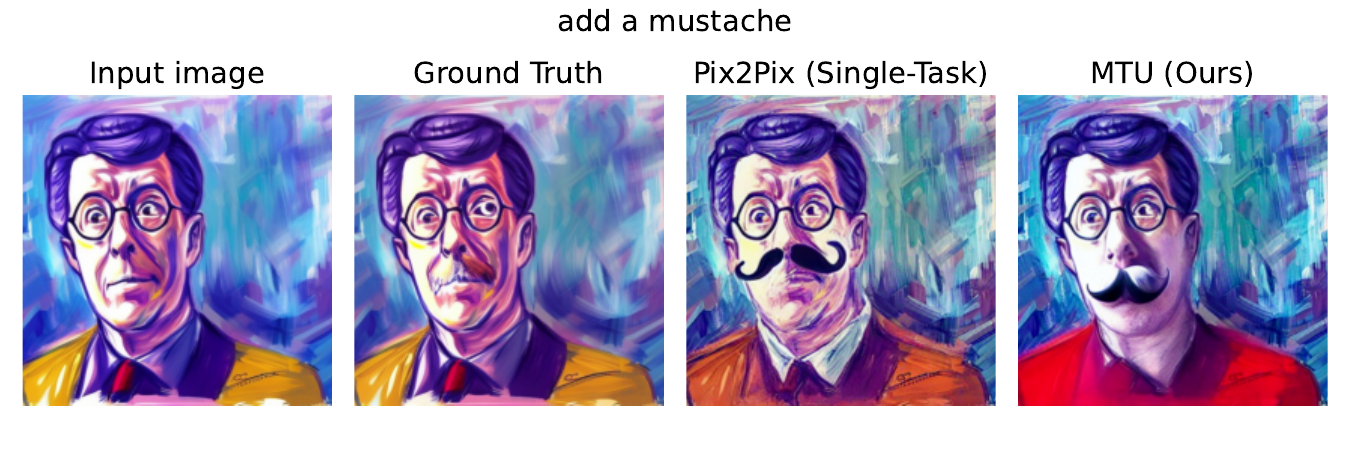}\put (210,65) {\small{SDv1.5}}
\end{overpic}
    }\end{subcaptionbox}
    \begin{subcaptionbox}{}{
\includegraphics[trim={{0cm} {1cm} {0cm} {0cm}}, clip,width=0.45\textwidth]{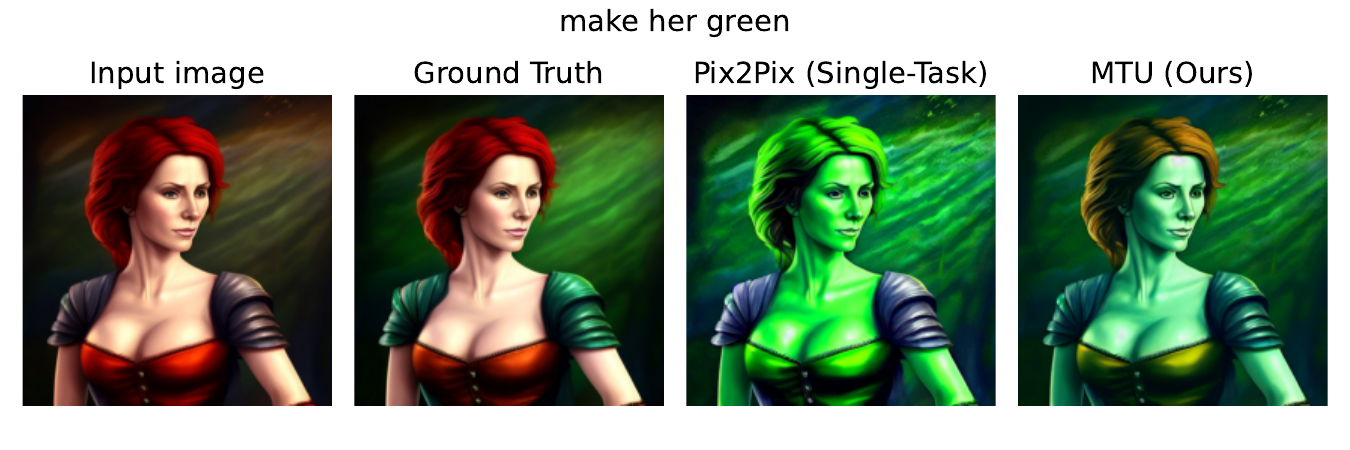}
    }\end{subcaptionbox}
        \begin{subcaptionbox}{}{
\includegraphics[trim={{0cm} {1cm} {0cm} {0cm}}, clip,width=0.45\textwidth]{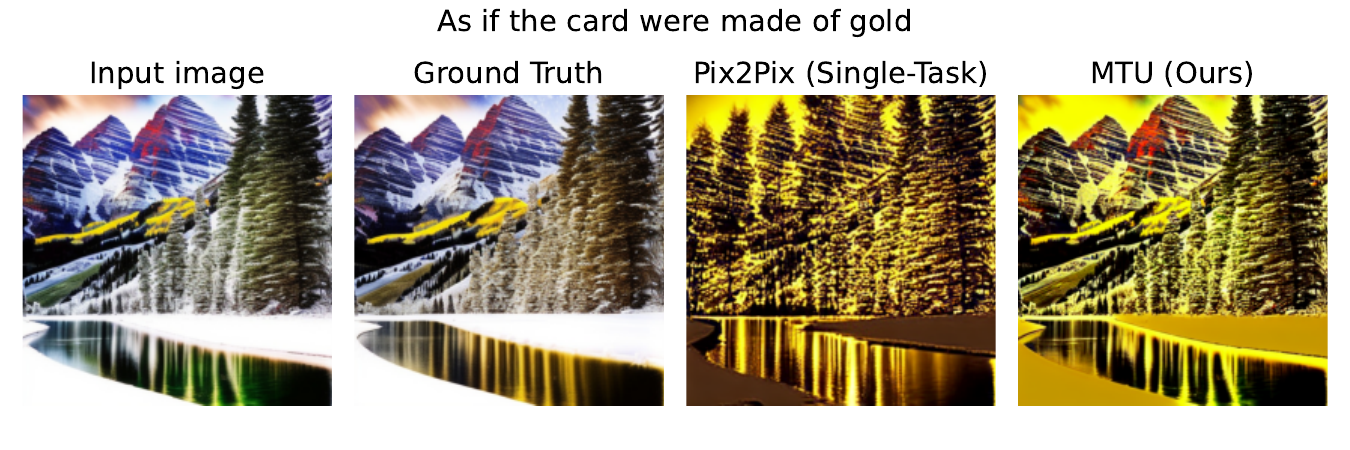}
    }\end{subcaptionbox}
    \begin{subcaptionbox}{}{
\includegraphics[trim={{0cm} {1cm} {0cm} {0cm}}, clip,width=0.45\textwidth]{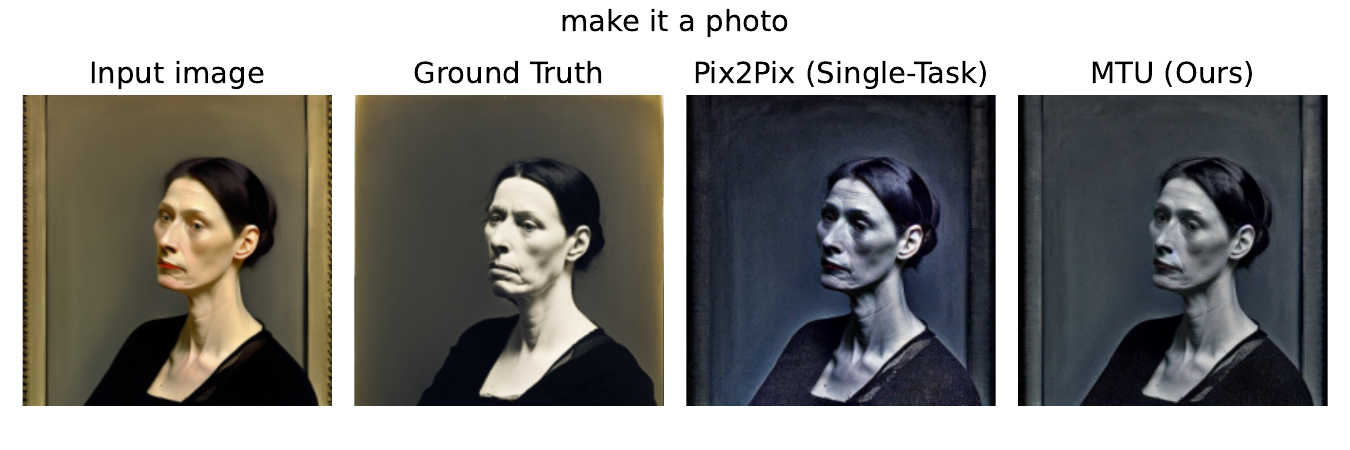}
    }\end{subcaptionbox}
    \begin{subcaptionbox}{}{
\includegraphics[trim={{0cm} {1cm} {0cm} {0cm}}, clip,width=0.45\textwidth]{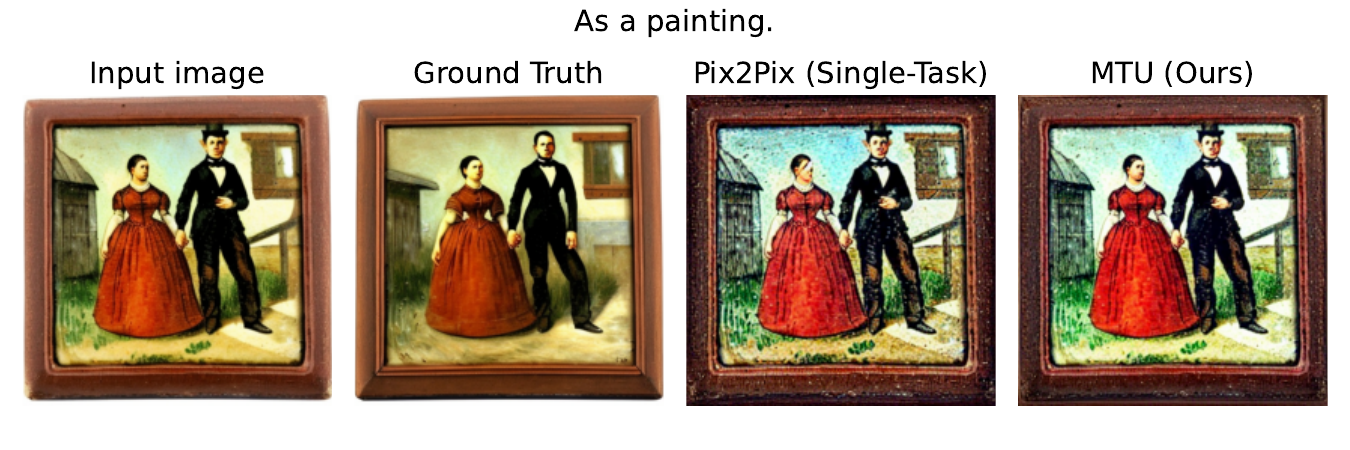}
\begin{subcaptionbox}{}{
\includegraphics[trim={{0cm} {1cm} {0cm} {0cm}}, clip,width=0.45\textwidth]{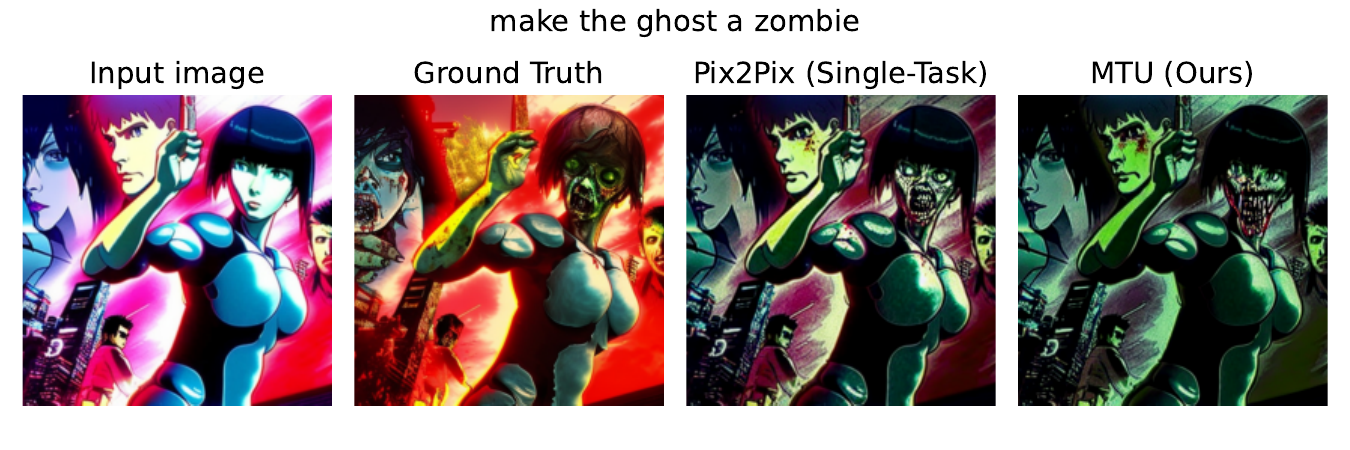}
    }\end{subcaptionbox}
    }\end{subcaptionbox}
    \begin{subcaptionbox}{}{
\begin{overpic}[trim={{0cm} {1cm} {0cm} {0cm}}, clip,width=0.45\textwidth]{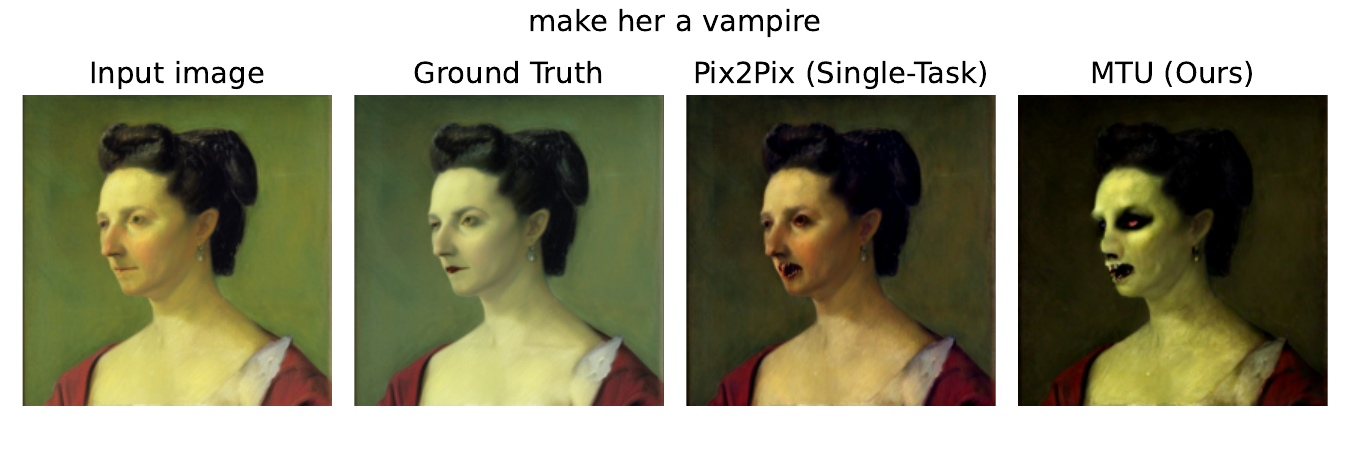}\put (210,65) {\small{SDXL}}
\end{overpic}
    }\end{subcaptionbox}
    \begin{subcaptionbox}{}{
\includegraphics[trim={{0cm} {1cm} {0cm} {0cm}}, clip,width=0.45\textwidth]{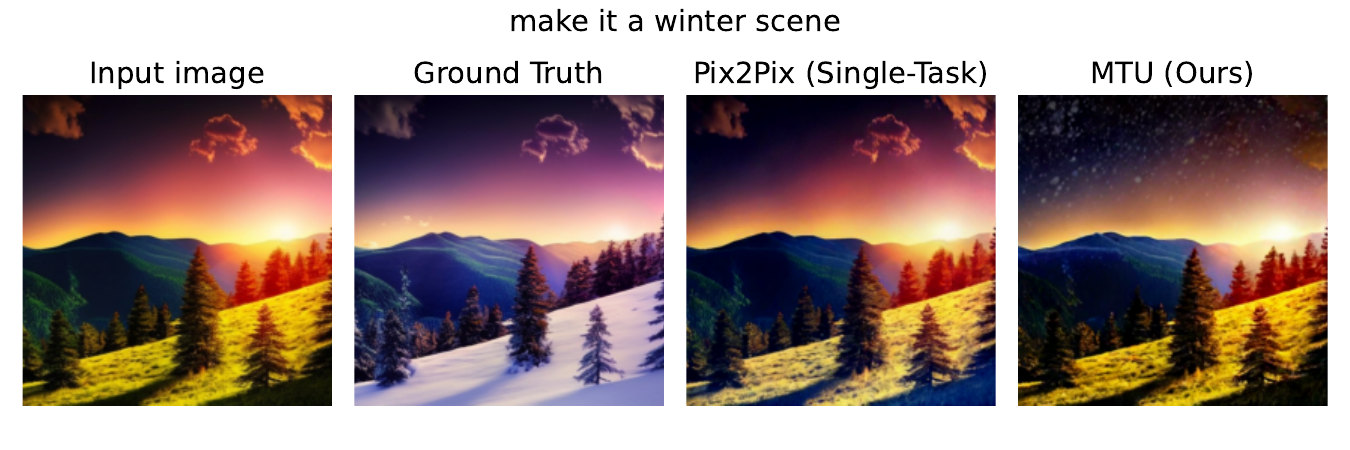}
    }\end{subcaptionbox}
        \begin{subcaptionbox}{}{
\includegraphics[trim={{0cm} {1cm} {0cm} {0cm}}, clip,width=0.45\textwidth]{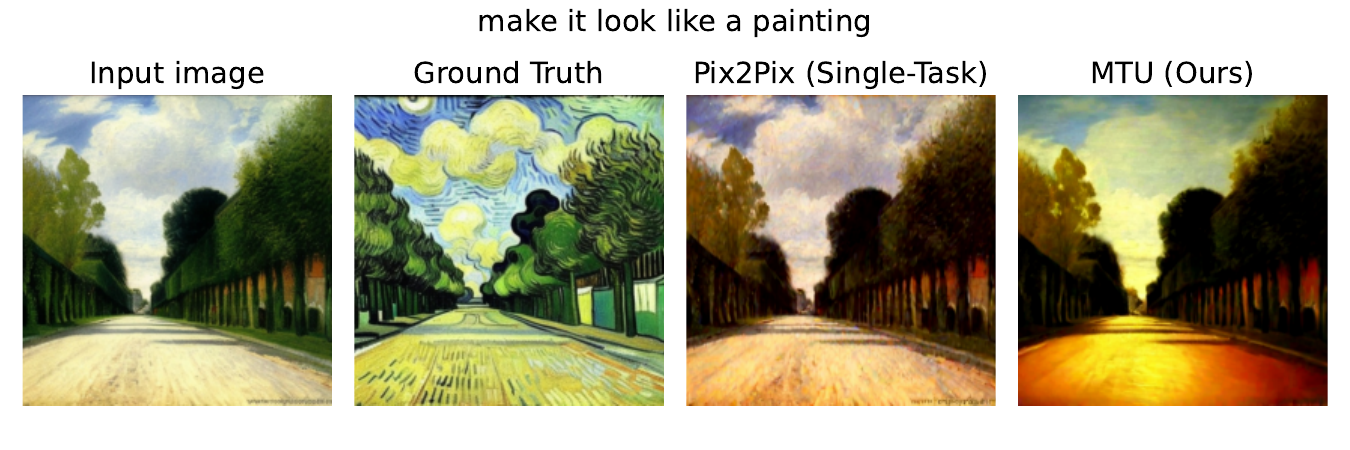}
    }\end{subcaptionbox}
    \begin{subcaptionbox}{}{
\includegraphics[trim={{0cm} {1cm} {0cm} {0cm}}, clip,width=0.45\textwidth]{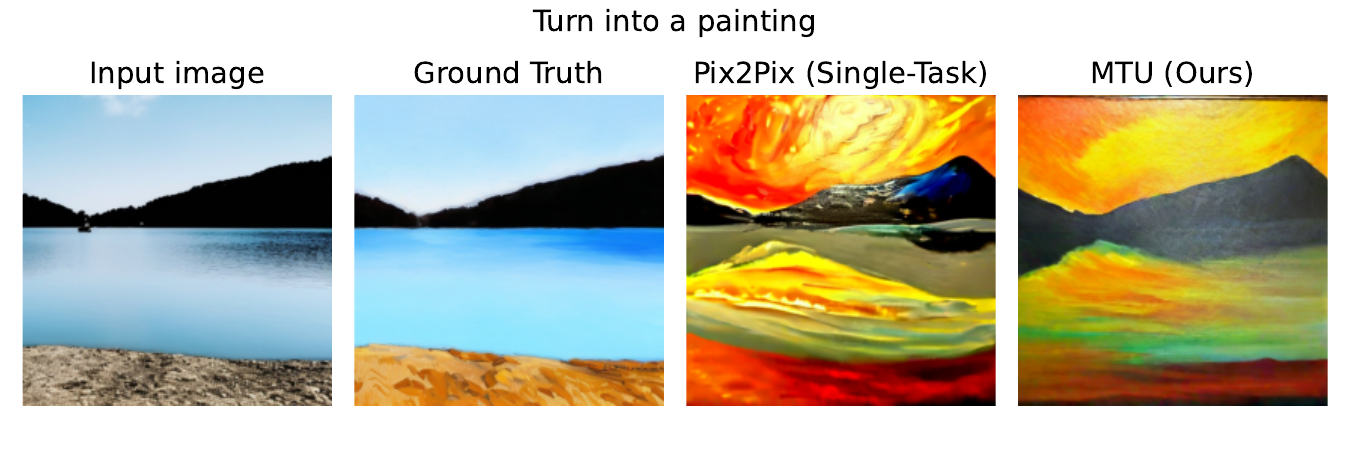}
    }\end{subcaptionbox}
    \begin{subcaptionbox}{}{
\includegraphics[trim={{0cm} {1cm} {0cm} {0cm}}, clip,width=0.45\textwidth]{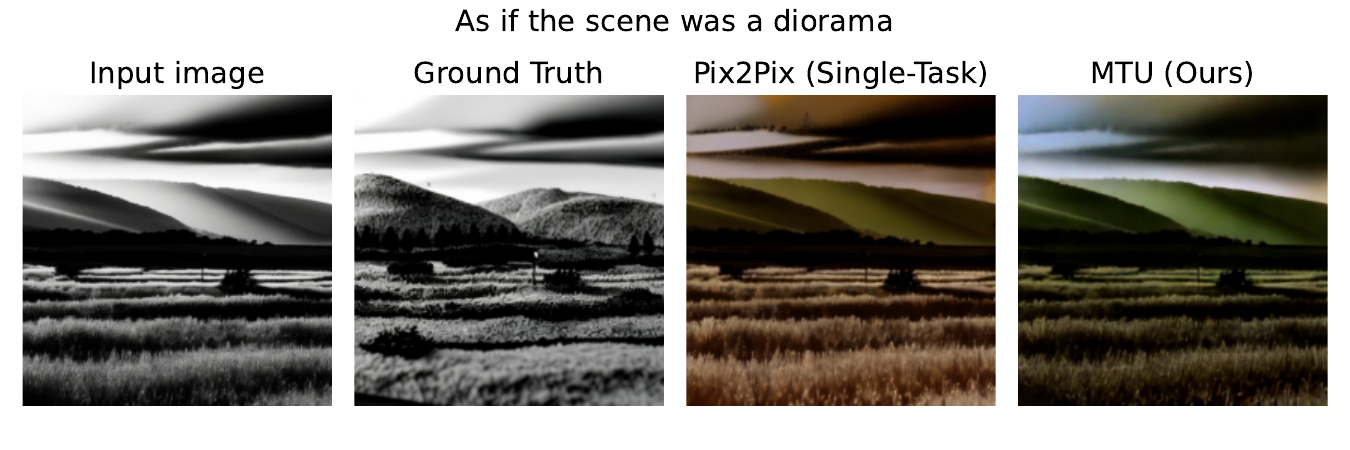}
    }\end{subcaptionbox}
    \begin{subcaptionbox}{}{
\includegraphics[trim={{0cm} {1cm} {0cm} {0cm}}, clip,width=0.45\textwidth]{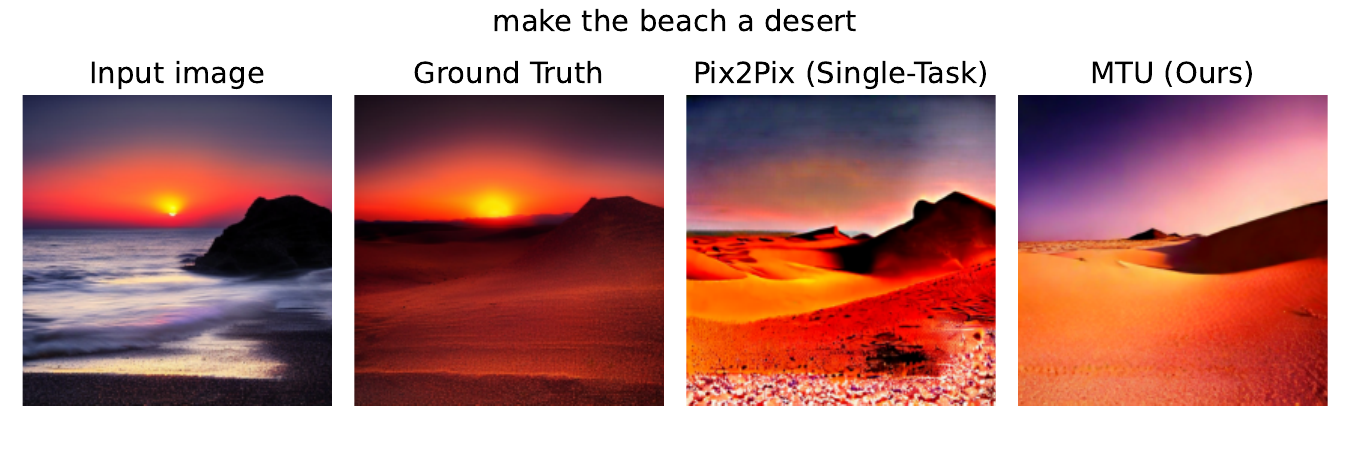}
    }\end{subcaptionbox}
    \caption{Qualitative results of MTU based on SDv1.5 and SDXL for Image Editing. In both models, our approach produces high-quality images with superior prompt adherence and faithful edits.}
    \label{fig:pics-ie}
\end{figure*}

\begin{figure*}[] 
 \captionsetup[subfigure]{labelformat=empty}
    \centering
    \begin{subcaptionbox}{}{
\begin{overpic}[trim={{0cm} {1cm} {0cm} {0cm}}, clip,width=0.45\textwidth]{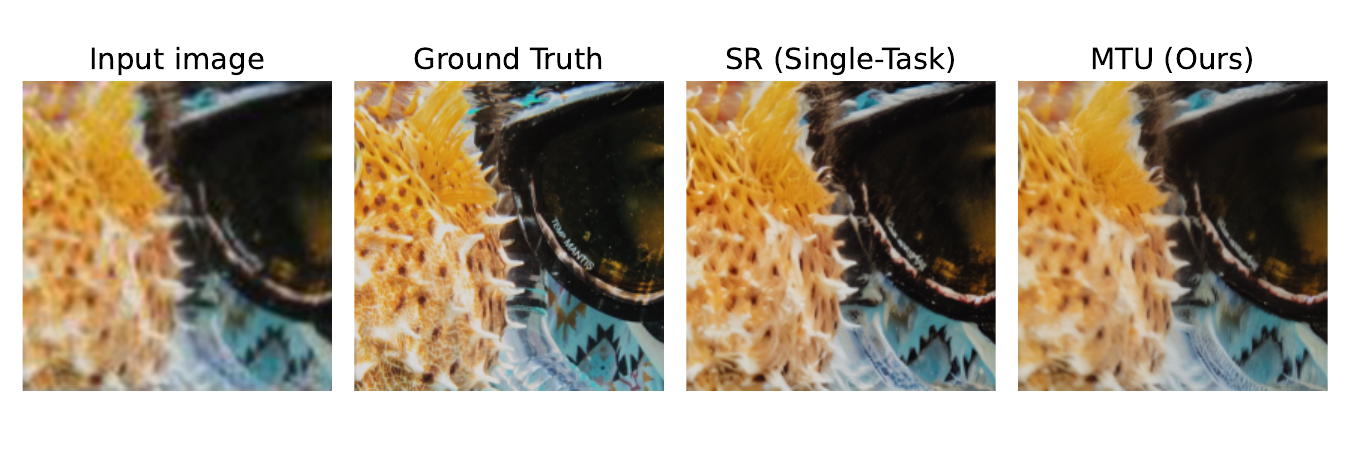}
    \put (210,65) {\small{SDv1.5}}
\end{overpic}}
\end{subcaptionbox}
    \begin{subcaptionbox}{}{
\includegraphics[trim={{0cm} {1cm} {0cm} {0cm}}, clip,width=0.45\textwidth]{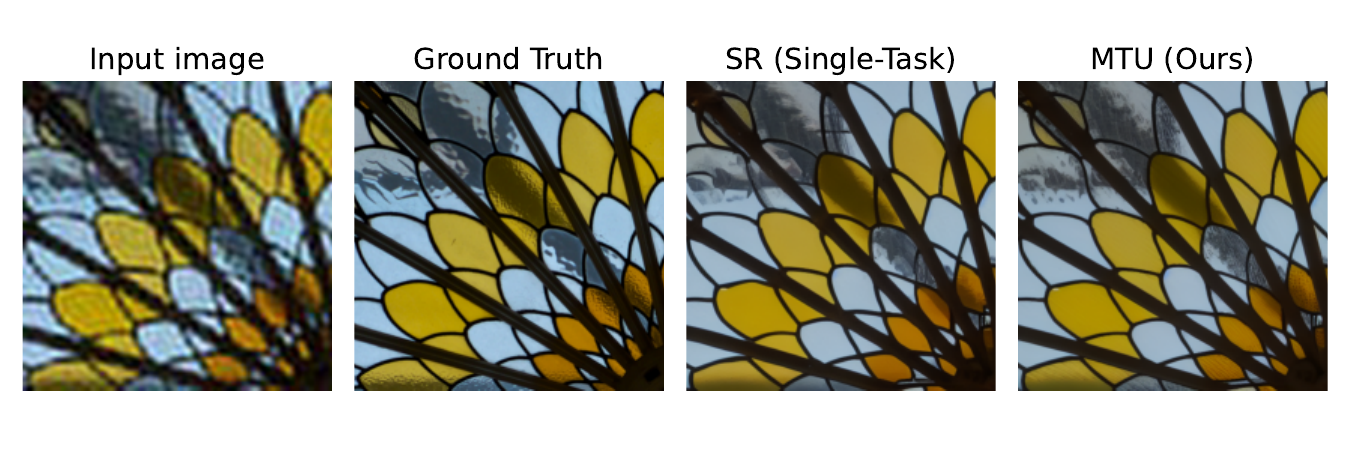}
    }\end{subcaptionbox}
        \begin{subcaptionbox}{}{
\includegraphics[trim={{0cm} {1cm} {0cm} {0cm}}, clip,width=0.45\textwidth]{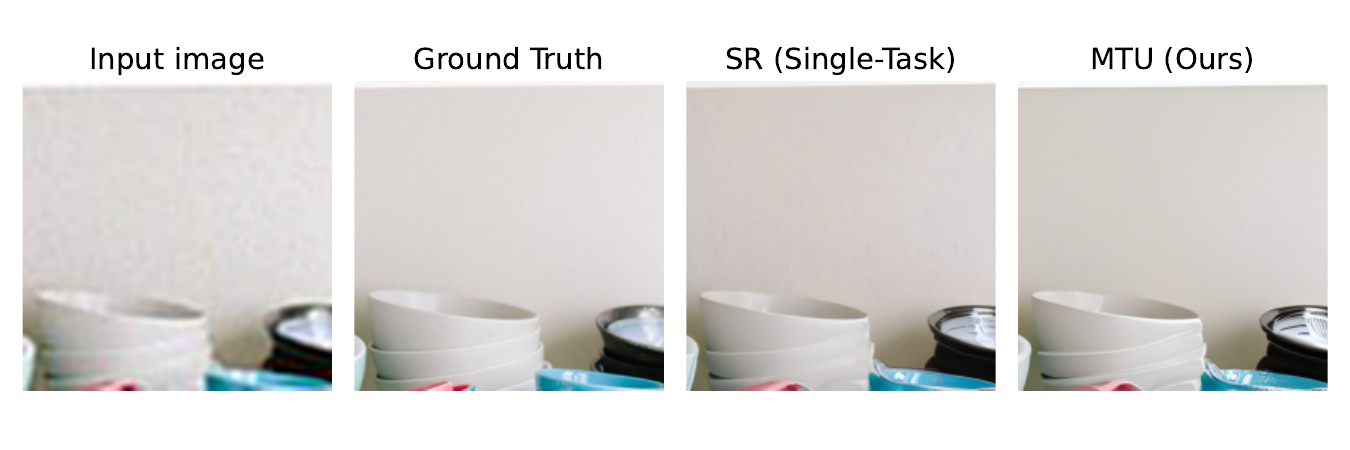}
    }\end{subcaptionbox}
    \begin{subcaptionbox}{}{
\includegraphics[trim={{0cm} {1cm} {0cm} {0cm}}, clip,width=0.45\textwidth]{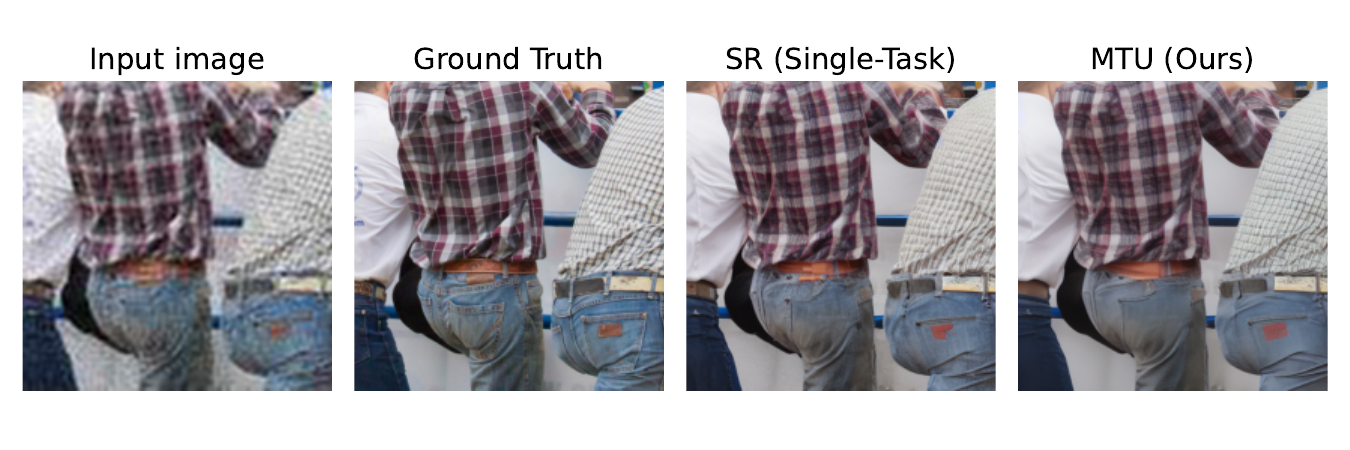}
    }\end{subcaptionbox}
     \begin{subcaptionbox}{}{
\includegraphics[trim={{0cm} {1cm} {0cm} {0cm}}, clip,width=0.45\textwidth]{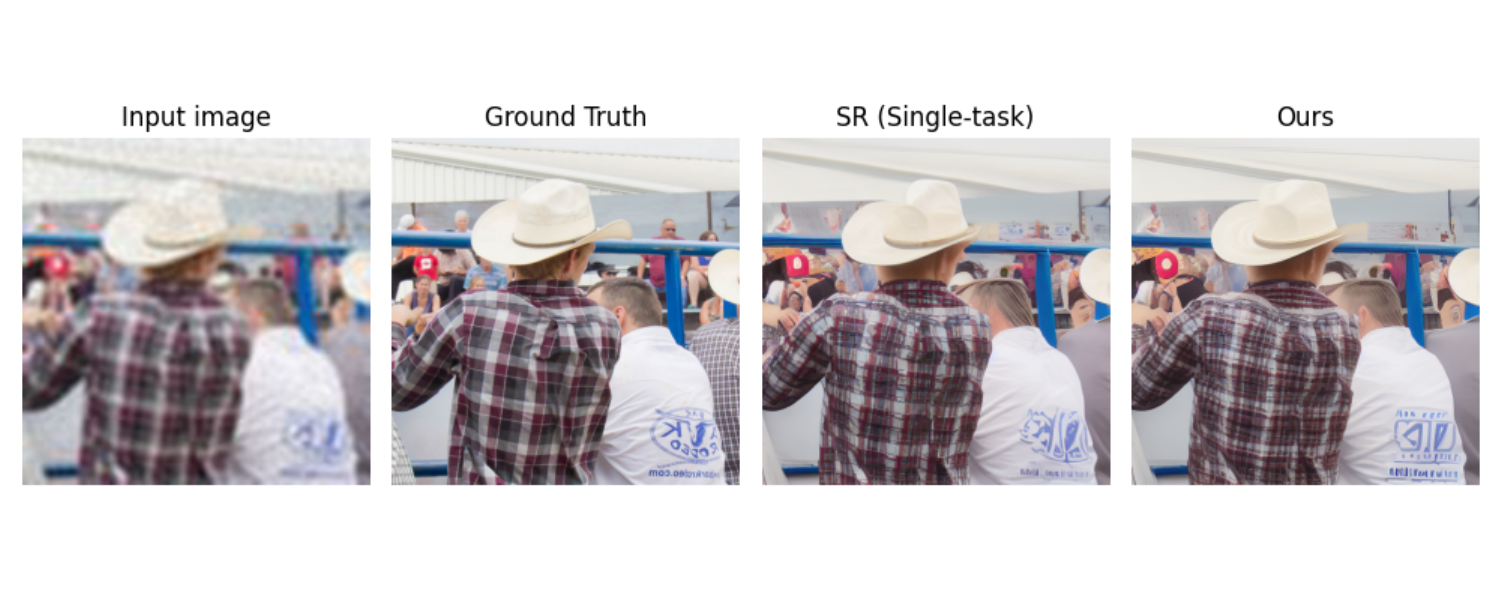}
    }\end{subcaptionbox}
    \begin{subcaptionbox}{}{
\includegraphics[trim={{0cm} {1cm} {0cm} {0cm}}, clip,width=0.45\textwidth]{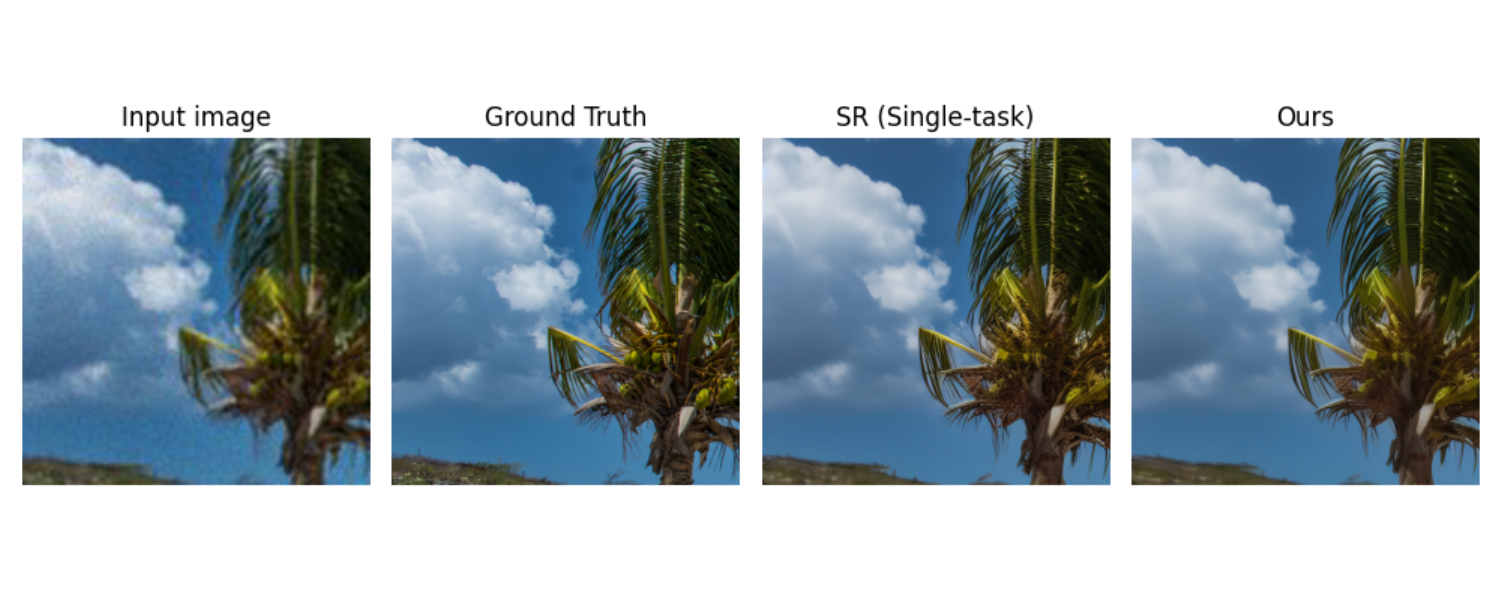}
    }\end{subcaptionbox}
      \begin{subcaptionbox}{}{
\begin{overpic}[trim={{0cm} {1cm} {0cm} {0cm}}, clip,width=0.45\textwidth]{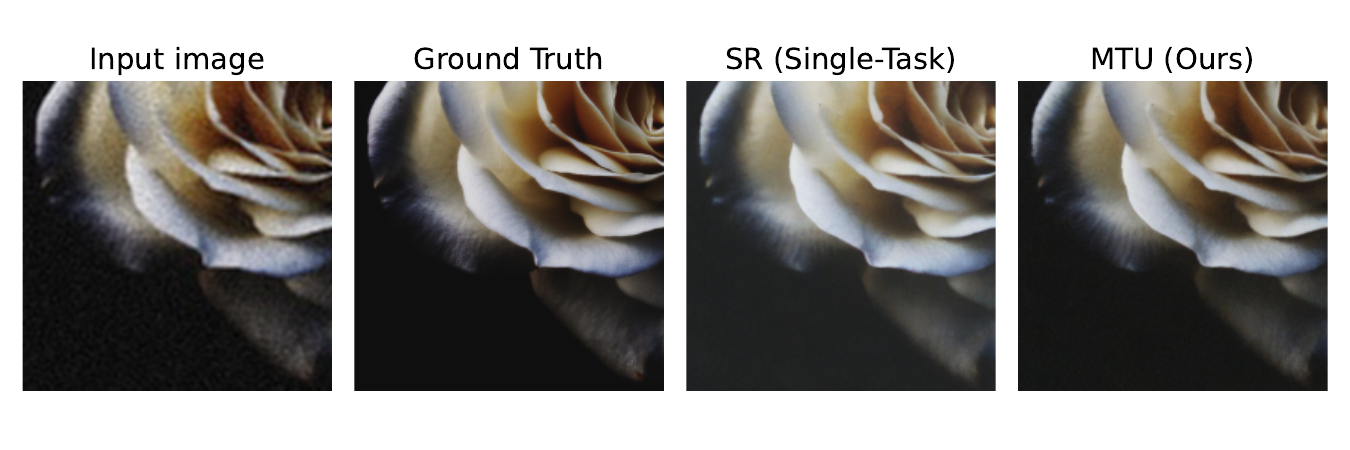}\put (210,65) {\small{SDXL}}
\end{overpic}
    }\end{subcaptionbox}
    \begin{subcaptionbox}{}{
\includegraphics[trim={{0cm} {1cm} {0cm} {0cm}}, clip,width=0.45\textwidth]{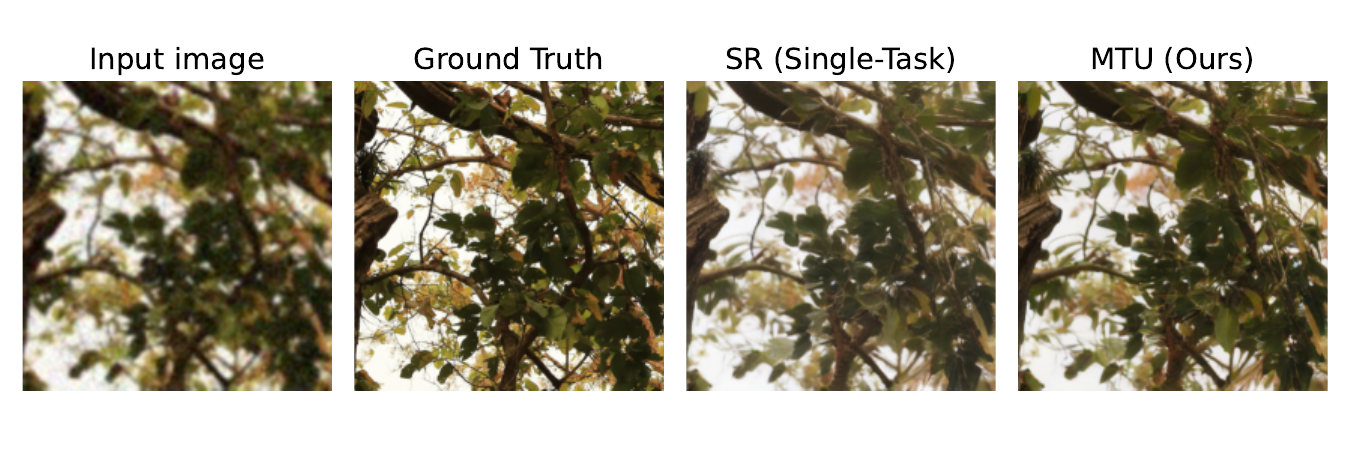}
    }\end{subcaptionbox}
        \begin{subcaptionbox}{}{
\includegraphics[trim={{0cm} {1cm} {0cm} {0cm}}, clip,width=0.45\textwidth]{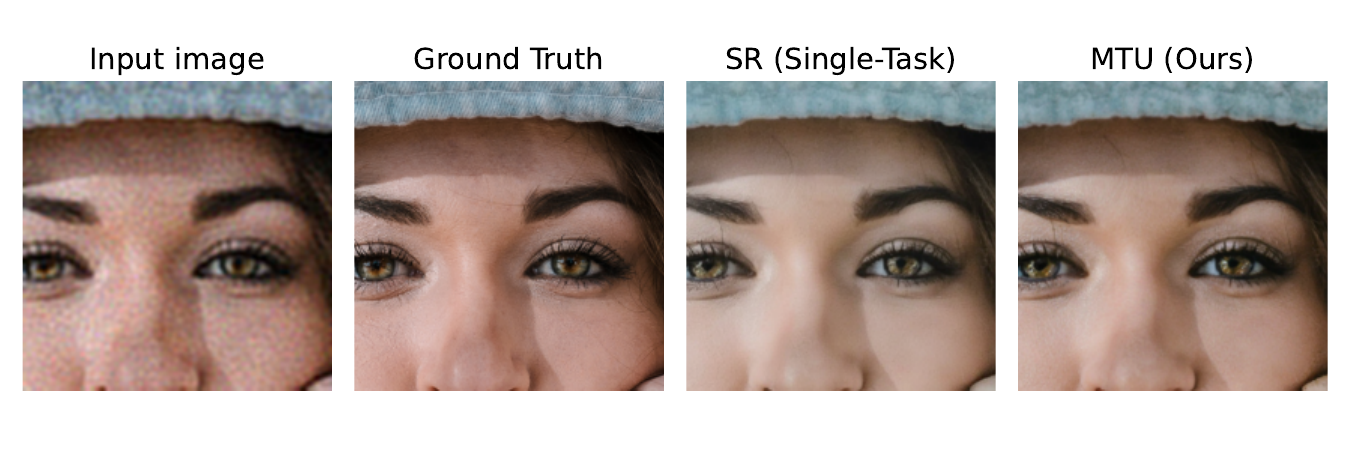}
    }\end{subcaptionbox}
    \begin{subcaptionbox}{}{
\includegraphics[trim={{0cm} {1cm} {0cm} {0cm}}, clip,width=0.45\textwidth]{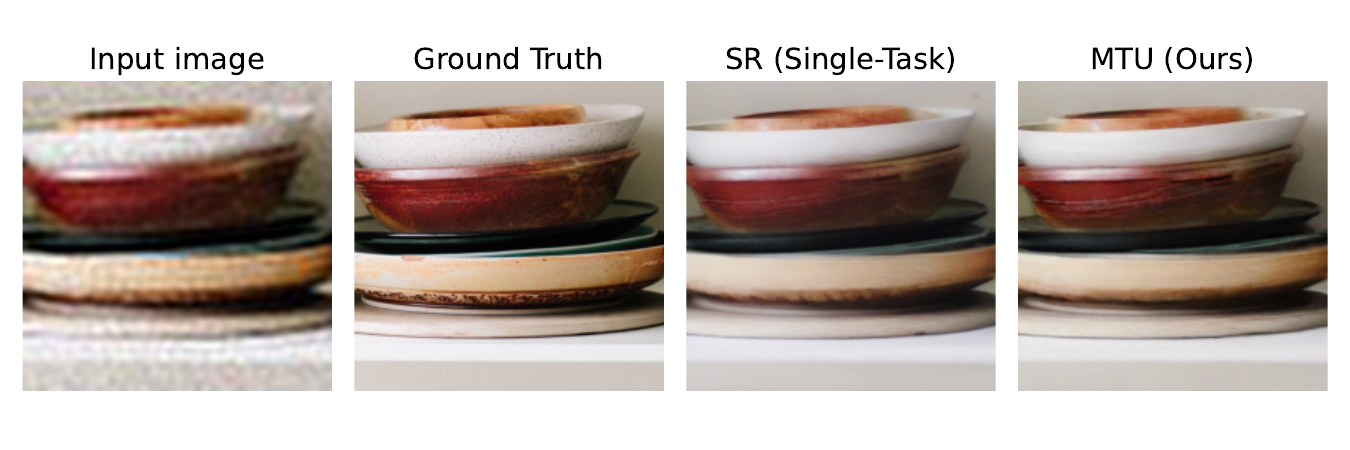}
    }\end{subcaptionbox}
    \begin{subcaptionbox}{}{
\includegraphics[trim={{0cm} {1cm} {0cm} {0cm}}, clip,width=0.45\textwidth]{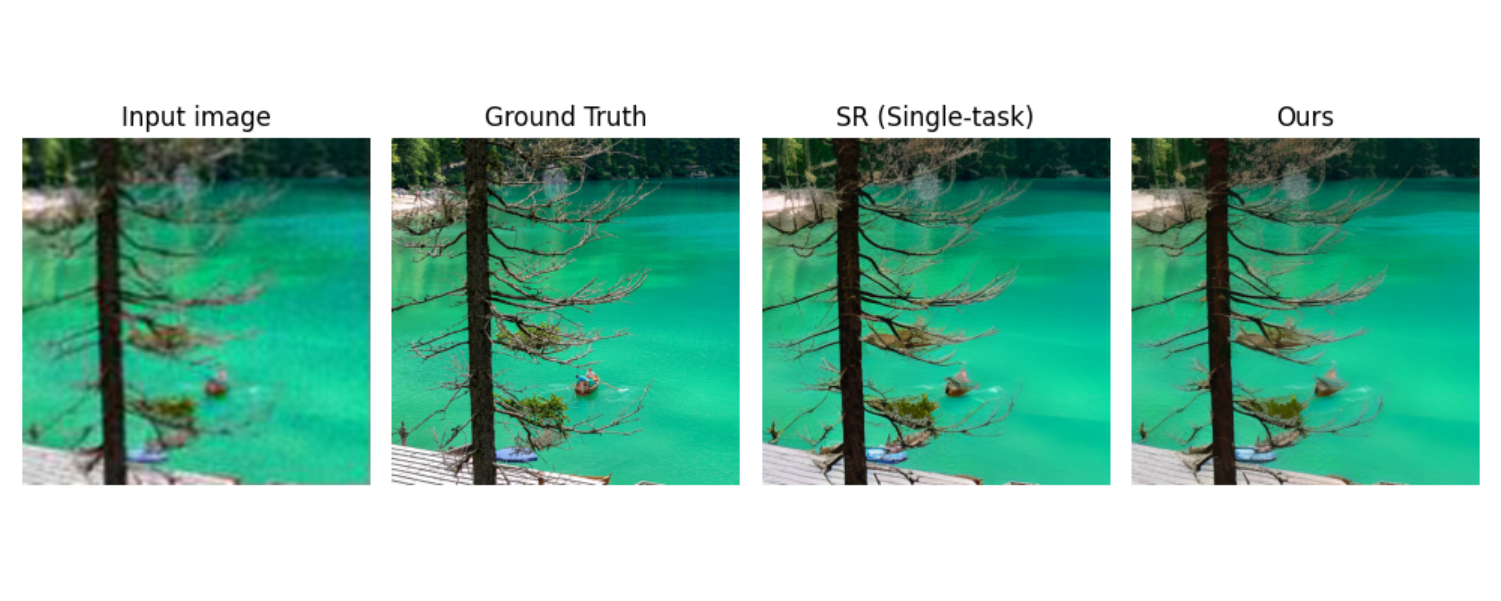}
    }\end{subcaptionbox}
    \begin{subcaptionbox}{}{
\includegraphics[trim={{0cm} {1cm} {0cm} {0cm}}, clip,width=0.45\textwidth]{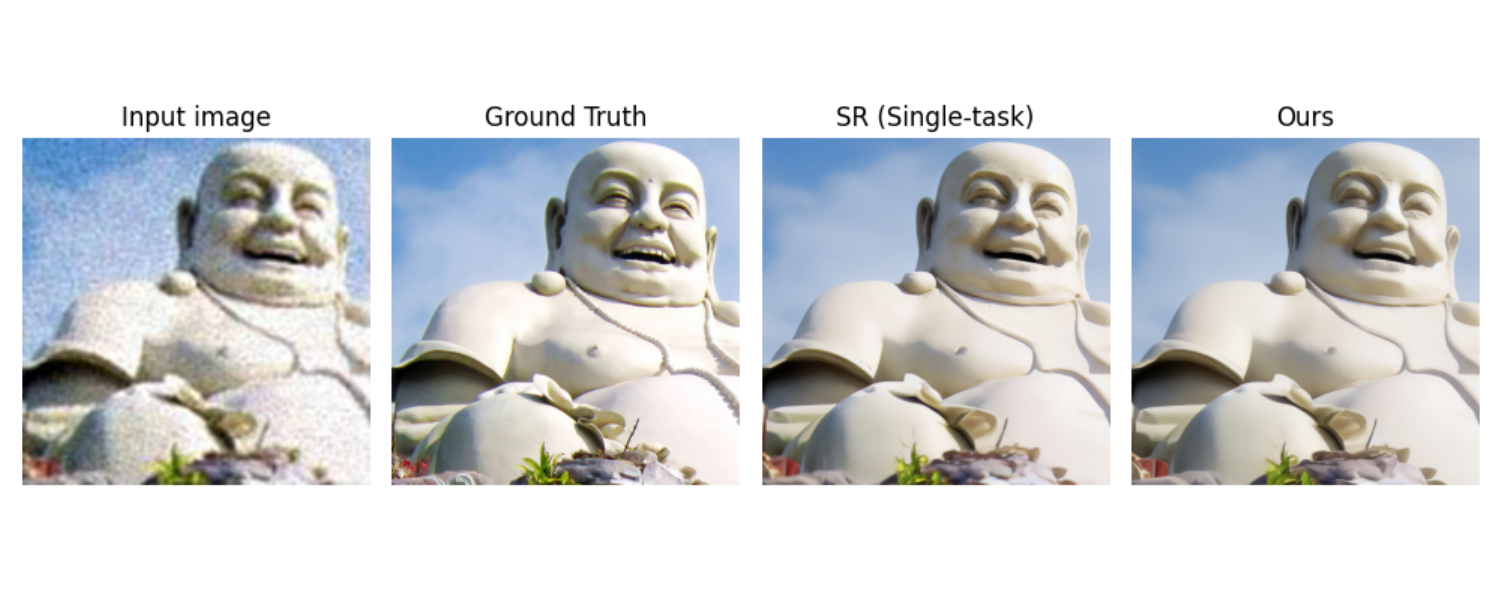}
    }\end{subcaptionbox}
    \caption{Qualitative results of MTU based on SDv1.5 and SDXL for Super Resolution. Our method effectively restores high-resolution images from low-resolution inputs that have been corrupted by image degradations, producing clearer and more detailed outputs.}
    \label{fig:pics-sr}
\end{figure*}

\begin{figure*}[] 
 \captionsetup[subfigure]{labelformat=empty}
    \centering
    \begin{subcaptionbox}{}{
\begin{overpic}[trim={{0cm} {1cm} {0cm} {0cm}}, clip,width=0.45\textwidth]{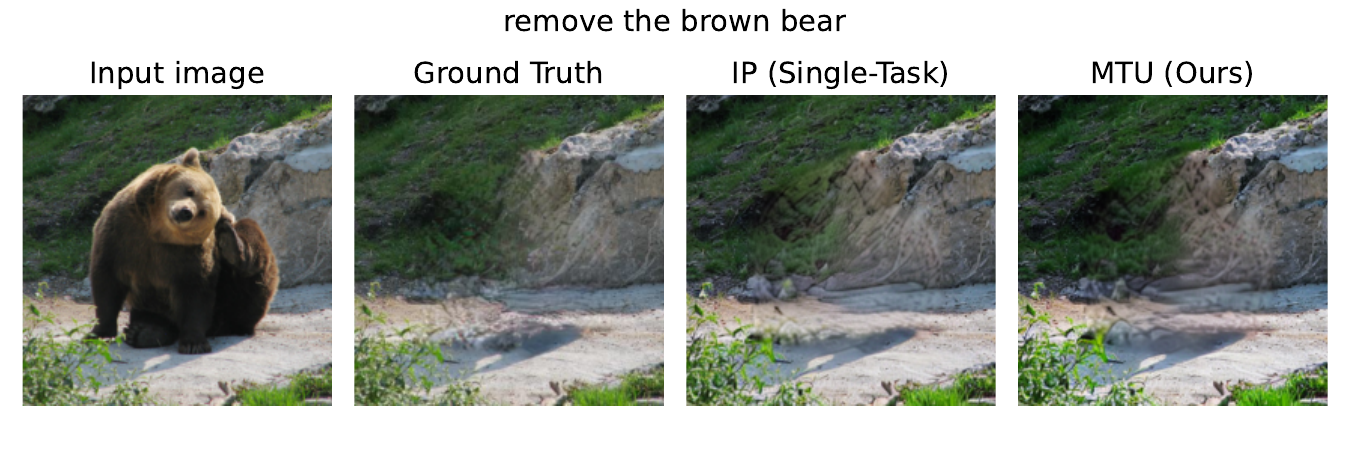}\put (210,65) {\small{SDv1.5}}
\end{overpic}
    }\end{subcaptionbox}
    \begin{subcaptionbox}{}{
\includegraphics[trim={{0cm} {1cm} {0cm} {0cm}}, clip,width=0.45\textwidth]{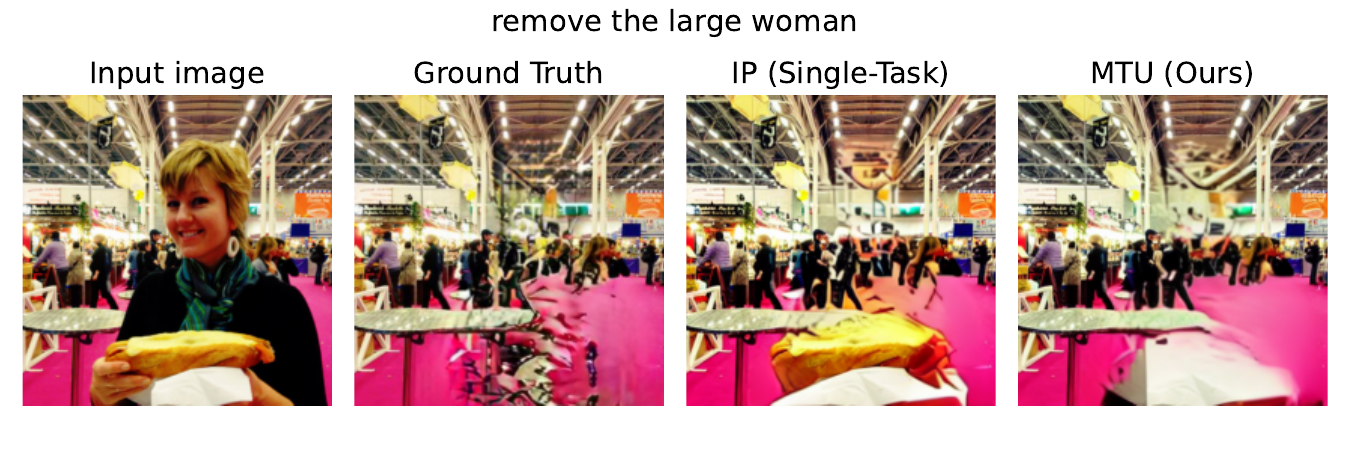}
    }\end{subcaptionbox}
        \begin{subcaptionbox}{}{
\includegraphics[trim={{0cm} {1cm} {0cm} {0cm}}, clip,width=0.45\textwidth]{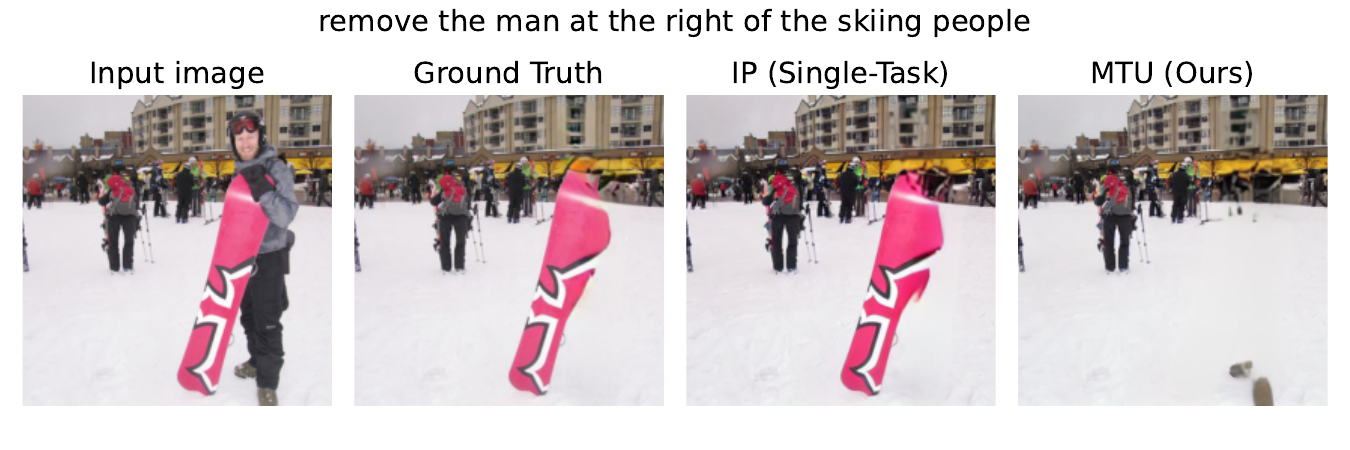}
    }\end{subcaptionbox}
    \begin{subcaptionbox}{}{
\includegraphics[trim={{0cm} {1cm} {0cm} {0cm}}, clip,width=0.45\textwidth]{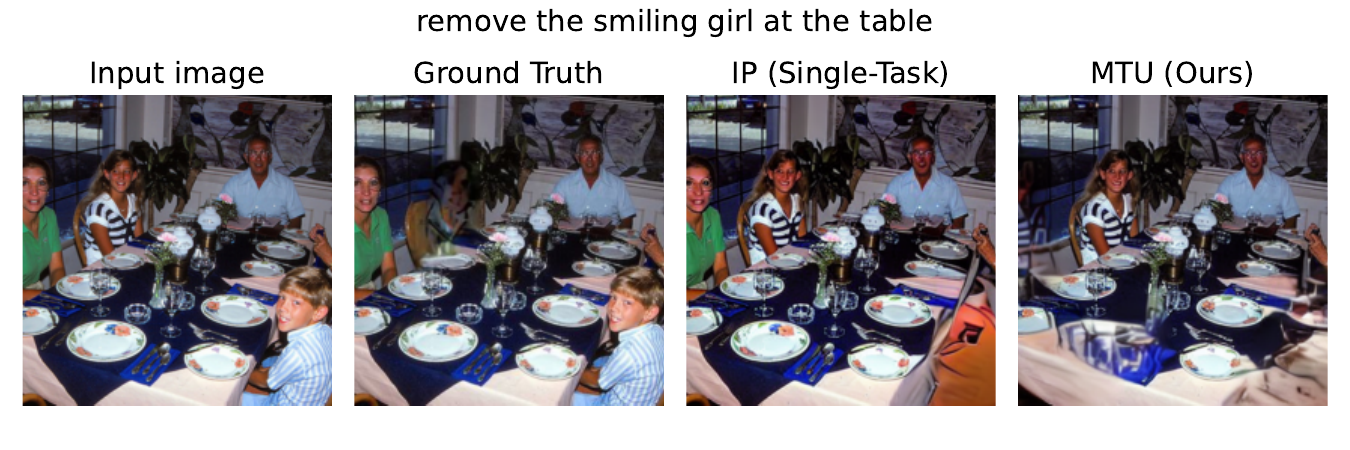}
    }\end{subcaptionbox}
       \begin{subcaptionbox}{}{
\includegraphics[trim={{0cm} {1cm} {0cm} {0cm}}, clip,width=0.45\textwidth]{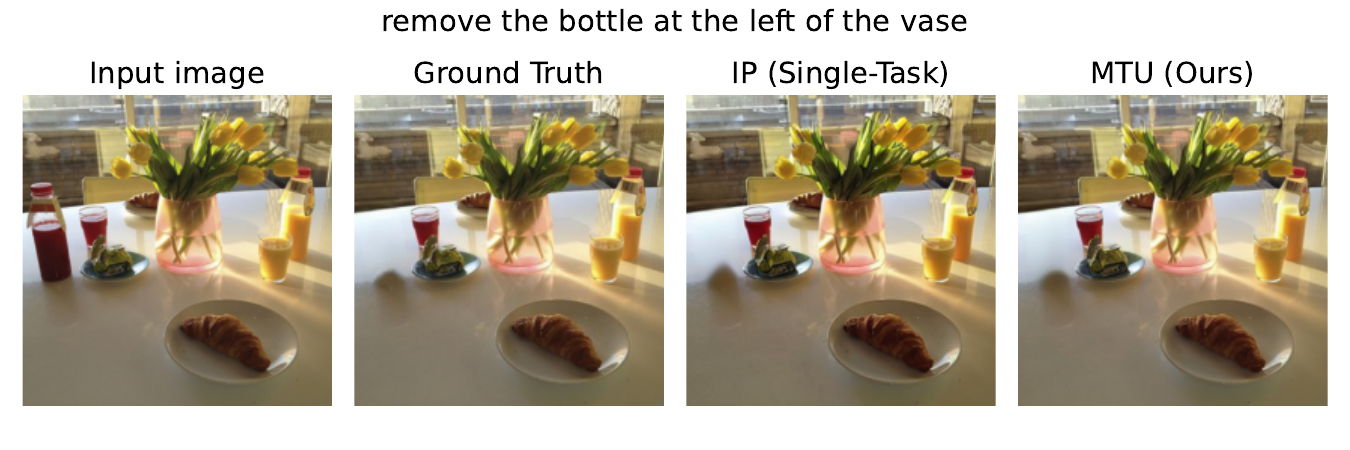}
    }\end{subcaptionbox}
    \begin{subcaptionbox}{}{
\includegraphics[trim={{0cm} {1cm} {0cm} {0cm}}, clip,width=0.45\textwidth]{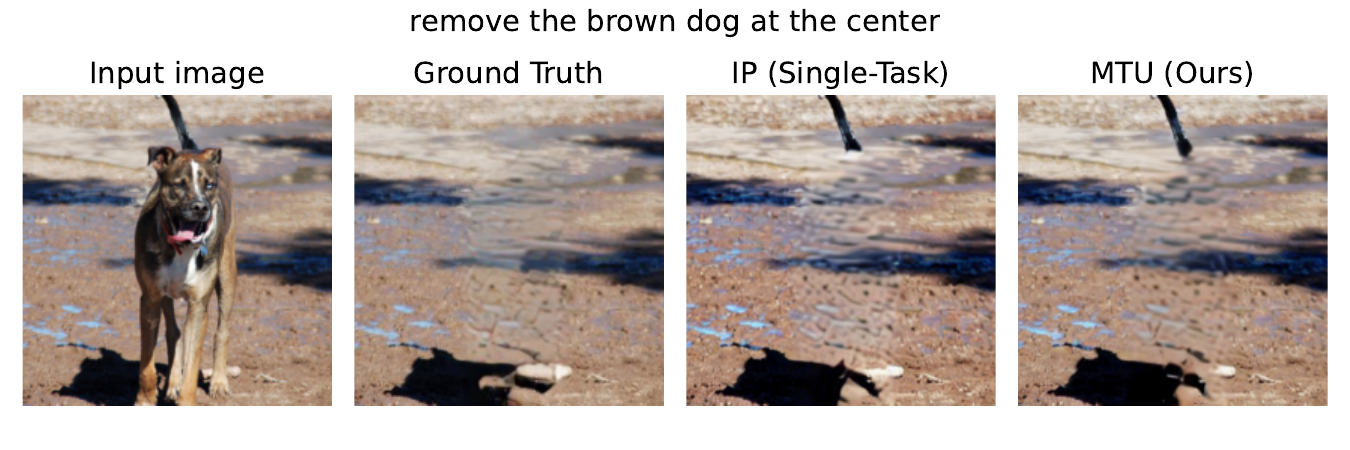}
    }\end{subcaptionbox}
    \begin{subcaptionbox}{}{
\begin{overpic}[trim={{0cm} {1cm} {0cm} {0cm}}, clip,width=0.45\textwidth]{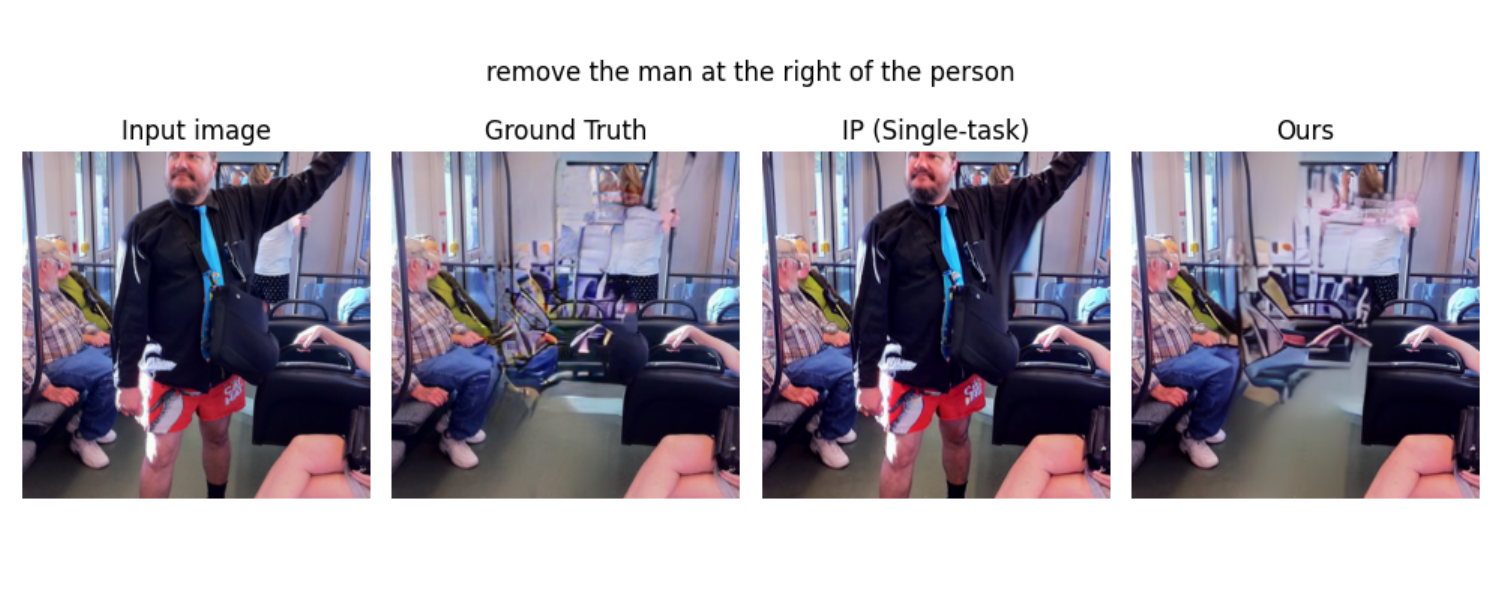}\put (210,80) {\small{SDXL}} \end{overpic}}
\end{subcaptionbox}
    \begin{subcaptionbox}{}{
\includegraphics[trim={{0cm} {1cm} {0cm} {0cm}}, clip,width=0.45\textwidth]{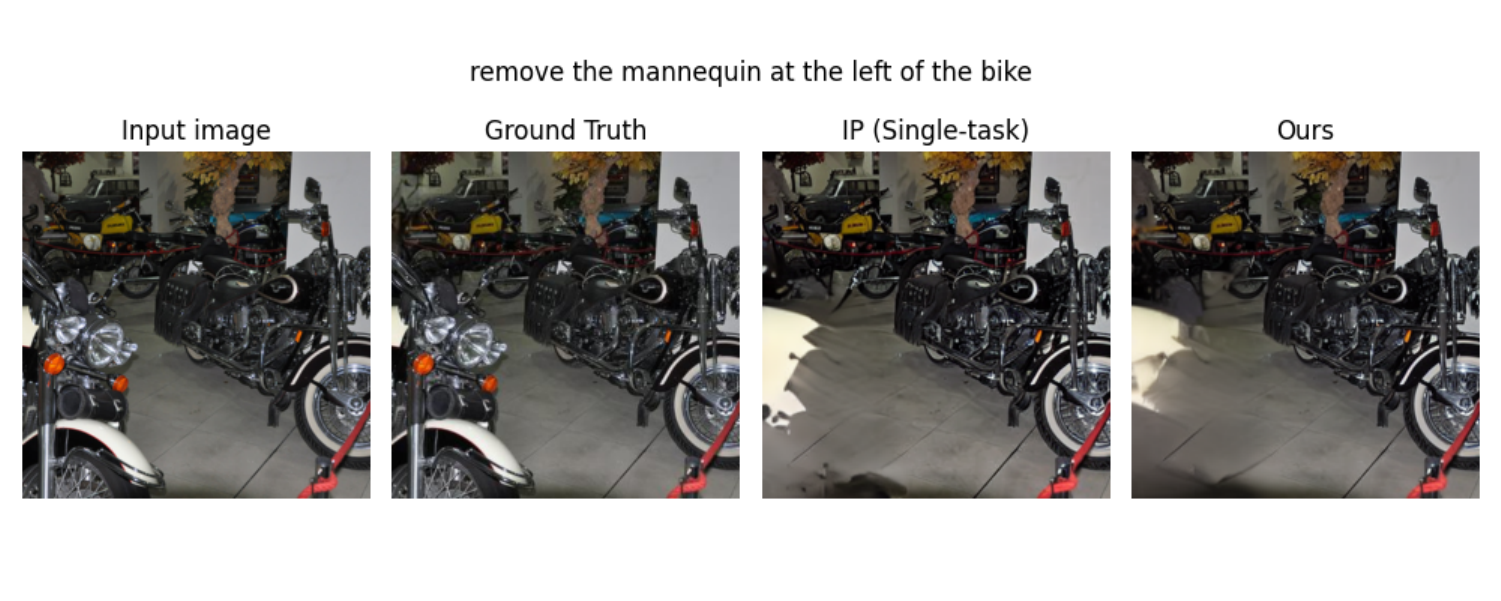}
    }\end{subcaptionbox}
        \begin{subcaptionbox}{}{
\includegraphics[trim={{0cm} {1cm} {0cm} {0cm}}, clip,width=0.45\textwidth]{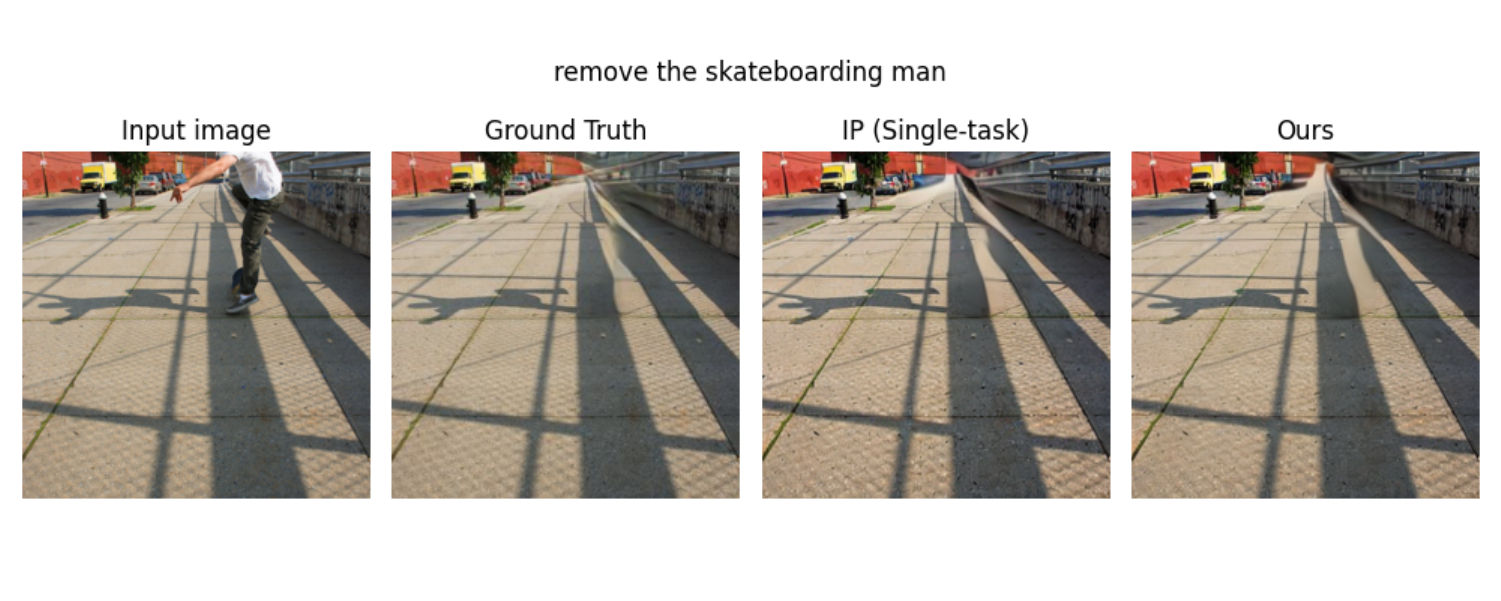}
    }\end{subcaptionbox}
    \begin{subcaptionbox}{}{
\includegraphics[trim={{0cm} {1cm} {0cm} {0cm}}, clip,width=0.45\textwidth]{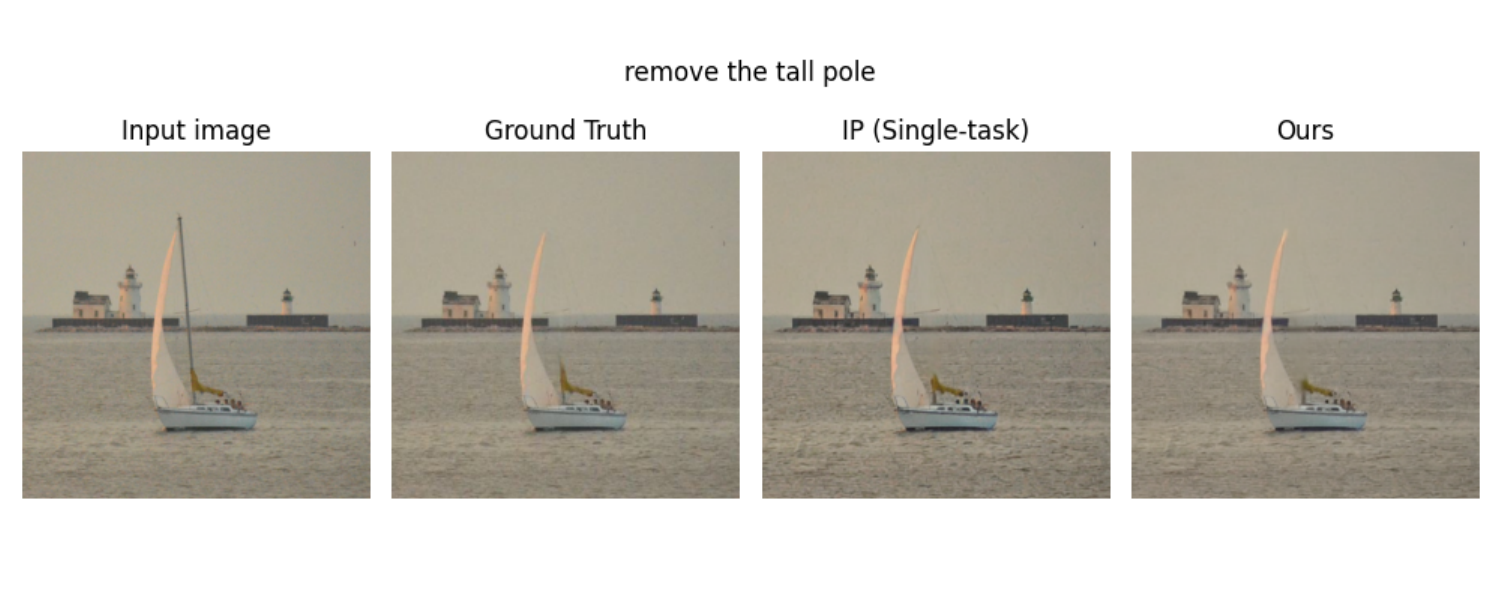}
    }\end{subcaptionbox}
      \begin{subcaptionbox}{}{
\includegraphics[trim={{0cm} {1cm} {0cm} {0cm}}, clip,width=0.45\textwidth]{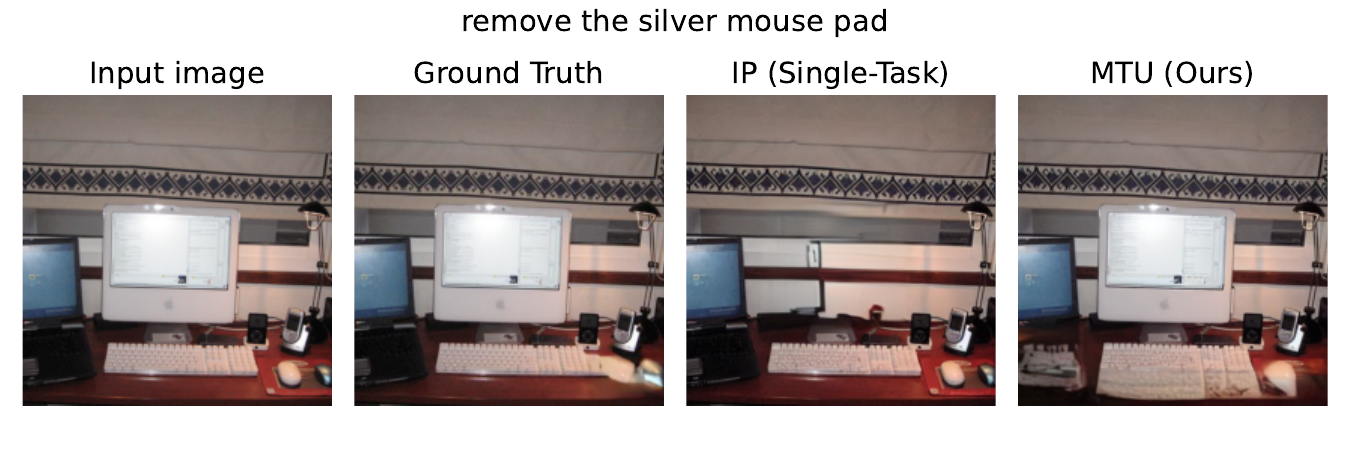}
    }\end{subcaptionbox}
    \begin{subcaptionbox}{}{
\includegraphics[trim={{0cm} {1cm} {0cm} {0cm}}, clip,width=0.45\textwidth]{figs/results/sd15_results/gqa_inpaint_remove_large_woman_5.pdf}
    }\end{subcaptionbox}
    \caption{Qualitative results of MTU based on SDv1.5 and SDXL for Image Inpainting. Our method demonstrates better object removal,  generating clean inpainted images.}
    \label{fig:pics-inpaint}
\end{figure*}

We provide more qualitative results for each of the tasks considered in the paper from Figures \ref{fig:pics-t2i}, \ref{fig:pics-ie}, \ref{fig:pics-sr}, and \ref{fig:pics-inpaint}.

\begin{figure}
    \centering
    \includegraphics[width=0.8\linewidth]{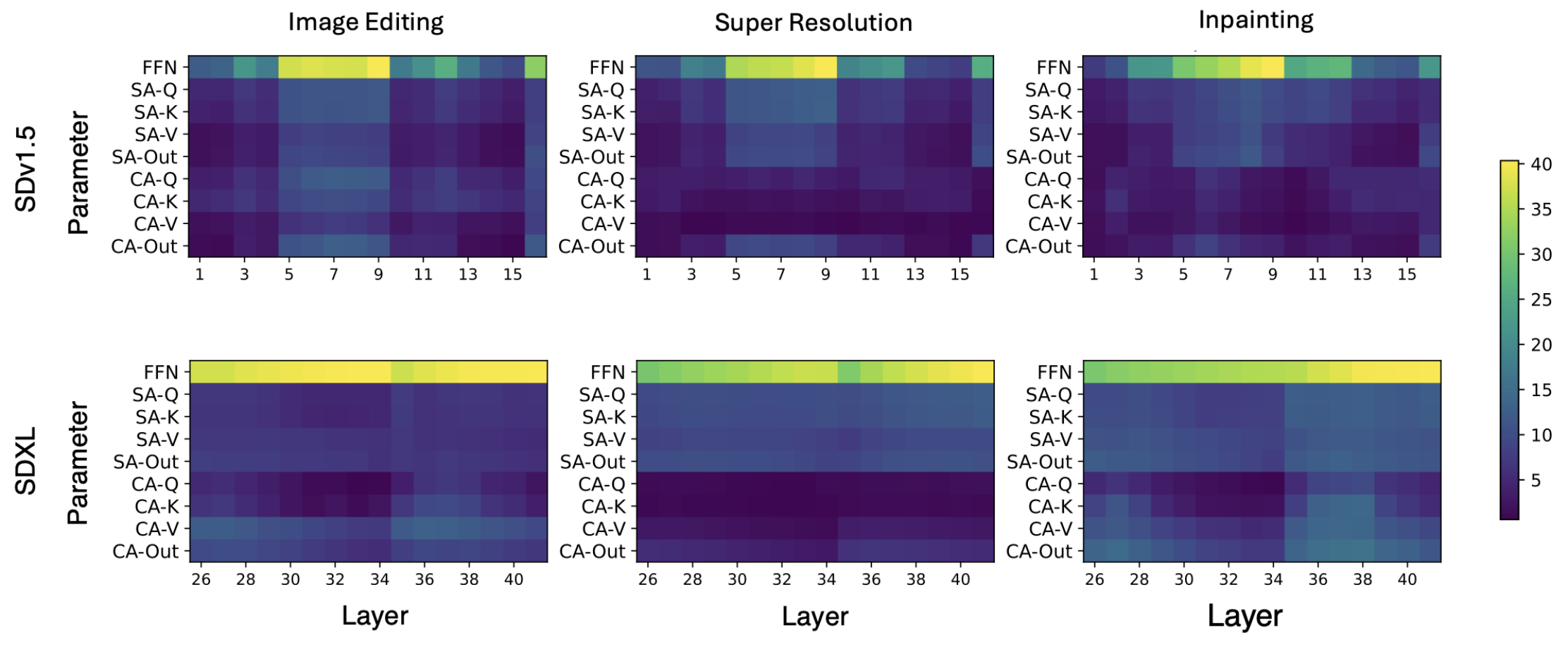}
    \caption{Distances between fine-tuned weights $\theta_{f}^{\tau}$ and pre-trained initialization $\theta_p$ for SDv1.5 (top) and SDXL (bottom) across Image Editing, Super-Resolution, and Inpainting. Here x-axis corresponds to the Layer index and the y-axis corresponds to the distance between the parameters. For all tasks, FFNs exhibit the highest deviation from initialization, highlighting their crucial role in adapting to downstream tasks. In SDXL, we focus only on the middle layers, where this deviation is most pronounced.}
    \label{fig:sd-distances}
\end{figure}



\end{document}